%% file: Camera_ready.tex
\pdfoutput=1

\documentclass[11pt]{article}

\usepackage[]{coling}
\usepackage{todonotes}
\usepackage{times}
\usepackage{latexsym}
\usepackage{listings}
\usepackage{blindtext}
\usepackage{tcolorbox}
\usepackage{graphicx}
\usepackage{multirow}
\usepackage{multicol}
\usepackage{subcaption}
\usepackage{lscape}
\usepackage{sidecap}
\usepackage{booktabs}
\usepackage{soul}
\usepackage{array}
\usepackage{arydshln}

\newcolumntype{C}[1]{>{\centering\arraybackslash\hspace{0pt}}p{#1}}

\usepackage[T1]{fontenc}

\usepackage[utf8]{inputenc}

\usepackage{microtype}

\usepackage{inconsolata}

\usepackage{graphicx}
\usepackage{enumitem}
\newcommand*\samethanks[1][\value{footnote}]{\footnotemark[#1]}
\title{QUENCH: Measuring the gap between Indic and Non-Indic Contextual General Reasoning in LLMs}

\author{Mohammad Aflah Khan\thanks{Equal Contribution}, Neemesh Yadav\samethanks, {\bf Sarah Masud}, {\bf Md. Shad Akhtar} \\
\texttt{\{aflah20082, neemesh20529, sarahm, shad.akhtar\}@iiitd.ac.in} \\
IIIT Delhi, India.}

\newcommand{\dataset}{\texttt{QUENCH}}

\begin{document}
\maketitle
\begin{abstract}
The rise of large language models (LLMs) has created a need for advanced benchmarking systems beyond traditional setups. To this end, we introduce \dataset, a novel text-based English \textbf{Qu}izzing B\textbf{ench}mark manually curated and transcribed from YouTube quiz videos. \dataset\ possesses masked entities and rationales for the LLMs to predict via generation. At the intersection of geographical context and common sense reasoning, \dataset\ helps assess world knowledge and deduction capabilities of LLMs via a zero-shot, open-domain quizzing setup. We perform an extensive evaluation on $7$ LLMs and $4$ metrics, investigating the influence of model size, prompting style, geographical context, and gold-labeled rationale generation. The benchmarking concludes with an error analysis to which the LLMs are prone.
\end{abstract}

\section{Introduction}
The ubiquitous rise of large language models (LLMs) has driven the need for diverse benchmarking and evaluation systems \cite{10.1145/3641289, peng2024survey}. Examining the logical reasoning and world knowledge capabilities of the LLMs \cite{qiao-etal-2023-reasoning, lu-etal-2023-survey} has been an active area of research. On one hand, world knowledge is primarily adjudged subject-wise (History, Law, STEM etc) via MMLU \cite{hendrycks2021measuring}, GSM-8k \cite{cobbe2021gsm8k} or via deductive multichoice (MCQ) question banks such as JEEBench \cite{arora-etal-2023-llms} or ScienceQA \cite{lu2022learn}. The subject-specific questions fail to capture the multi-themed nature of real-world knowledge and reasoning, which involves relating concepts from multiple different ``subject" areas. Furthermore, the MCQ setup already provides a plausible answer and restricts the answering domain to a closed-source one. Meanwhile, the commonsense understanding is broadly examined via the likes of pragmatism \cite{sravanthi2024pub}, syllogism \cite{wu-etal-2023-hence}, and truthfulness \cite{lin-etal-2022-truthfulqa} with minimal references to real-world historical events and entities. Parallelly, researchers have also been proving the efficacy of LLMs as evaluation tools, such as G-Eval \cite{geval}. 

\begin{table*}[ht!]
    \centering
    \resizebox{\textwidth}{!}
    {
    \begin{tabular}{lp{52em}l}
        \toprule
     \multirow{9}{*}{\rotatebox{90}{\bf Prompt}} 
     &
         \textbf{Quiz Prompt:} Consider yourself a participant in a quiz show where I am the quizmaster. I will ask you a question that can be from any general theme. You need to provide me with the correct answer. The question can have multiple variables to answer, and you need to provide me with the answer for variables X, Y, and Z. Hence, use the following format strictly in your response: `The answer is $<$X:~answer$>$.' You will lose points if you fail to follow the format. & \\
         & \textbf{Question:} \em The primary protagonist of \textbf{X}, one of China's classic epics, derived his literary name from items he carried and shared it with a rather famous mudra gesture. However, he is less well-known than \textbf{Y}, a secondary protagonist in the original text who was an allegory for the restless human mind. \textbf{Z} is the protagonist of one of the most famous manga ever, adapted to over 20 movies and spawned many memes thanks to its unique transformations. \textbf{Z} shares his name and characterization with \textbf{Y}. & \\ \cmidrule{2-2}
         \multirow{6}{*}{\rotatebox{90}{\bf Outputs}} & \textbf{Predicted entities:} $<$\textbf{X:} \textit{Journey to the west}$>$; 
         $<$\textbf{Y:} \textit{Sun Wukong}$>$; 
         $<$\textbf{Z:} \textit{Son Goku}$>$; & \\ 
         & \textbf{Rationale:}  X is Journey to the West, which features the protagonist Sun Wukong, who is known as the Monkey King. He carries a staff and is associated with the mudra gesture. Y is Sun Wukong, who is the Monkey King and secondary protagonist in Journey to the West. He represents the restless human mind. Z is Son Goku from the Dragon Ball series. He shares his name and monkey-like characterization with Sun Wukong, the Monkey King from Journey to the West. Dragon Ball has spawned over 20 movies and many memes. \\ \midrule \midrule

          \multirow{7}{*}{\rotatebox{90}{\bf Prompt}} 
     &
         \textbf{Quiz Prompt:} Consider yourself a participant in a quiz show where I am the quizmaster. I will ask you a question that can be from any general theme. You need to provide me with the correct answer. The question can have multiple variables to answer, and you need to provide me with the answer for variable X. Hence, use the following format strictly in your response: `The answer is $<$X:~answer$>$.' You will lose points if you fail to follow the format. & \\
         & \textbf{Question:} \em Karondi is a remote village near Jabalpur. \textbf{X} of India was moved to Karondi from Nagpur after the partition. \textbf{X} was established during the Great Trigonometrical Survey of India in 1907 in Nagpur. The project, basically, was to demarcate the British territories in India. & \\ \cmidrule{2-2}
         \multirow{3}{*}{\rotatebox{90}{\bf Outputs}} & \textbf{Predicted entities:} $<$\textbf{X:} \textit{Centre Point of India}$>$; &
         \\ 
         & \textbf{Rationale:} The Centre Point of India was moved post the partition due to a change in the boundaries of the country. It was established during the Great Trigonometrical Survey of India in 1907 in Nagpur. \\
         \bottomrule
    \end{tabular}}
    \caption{An overview of the quiz evaluation setup with zero-shot prompt (without CoT in this case) for \textit{predicting the masked entities (X, Y, Z)} in a quiz question taken verbatim from \dataset. The prompt and question are provided to an LLM, and the expected generations are to be the missing entities in the form $<$X: answer$>$. For the CoT setup, the above ``quiz prompt" is suffixed with the phrase `Let's think step by step.` Each question has the required masked entities (for instance, `X' or `X, Y, Z') in the text, and we simply refer to each of them via ``The question can have multiple variables to answer, and you need to provide me with the answer for variable <variable>".}
    \label{tab:quiz_examples}
\end{table*}

Another frequently overlooked issue with benchmarks is their tendency to be ``\textit{global}" or ``\textit{western-centric}" \cite{whose_opinions, durmus2024measuringrepresentationsubjectiveglobal}. For a language model to understand complex cultural queries, it must first be acquainted with the specific entities and concepts of those cultures. 
While efforts like \citet{seth-etal-2024-dosa} have been made, to the best of our knowledge, none present an open-world, clue-based guessing game where disjoint clues rely on linguistic hints or require combining multiple concepts to arrive at the answer. To address this gap, we propose a quiz trivia-based framework to probe LLMs' deductive reasoning skills across multiple themes and entities. Our work focuses on world knowledge and commonsense reasoning, with an emphasis on geographical context (Table \ref{tab:quiz_examples}). 

\textbf{Benchmark Dataset.} To support our investigation, we curate \dataset, a \textbf{Qu}izzing B\textbf{ench}mark. It is a novel text-only dataset obtained from manually annotated YouTube quiz videos in English. The primary entity to be predicted is hidden in the form of ``X'' and can occur multiple times in the question along with multiple entities as well (example \# 2 in Table \ref{tab:quiz_examples}). We also provide a manually curated free-text explanation/rationale for the correct answers. The non-trivial coreference resolutions across entities uniquely position \dataset\ to access the deduction capabilities of the LLMs with respect to the world and commonsense knowledge. 

\textbf{Benchmarking and Evaluation Setup.} We benchmark entity and rationale prediction for \dataset\ across seven different LLMs, including both closed and open-weight models. These models vary significantly in terms of the number of parameters, knowledge cutoff dates, context lengths, and pretraining characteristics. We employ the standard metrics -- BLEU, BERTScore, and ROUGE-L, as well as a GEval-based strategy to evaluate the performance under zero-shot prompting both with and without chain-of-thought (CoT) as outlined in Table \ref{tab:quiz_examples}. The Indic subset (example \# 1 in Table \ref{tab:quiz_examples}) in \dataset\ further allows for examining the Indian knowledge representation in the LLMs. 

\textbf{Observations.} Based on GEvals, our analysis reveals that zero-shot entity prediction accuracy improves from 72\% to 87\% when upgrading from GPT-3.5 to GPT-4. As expected, GPT-4 leads across all four metrics. Meanwhile, the open-weight LLaMA-3-70B performs on par with GPT-3.5 and stands out as a strong alternative, mainly due to its lower variability across subsets (12 points compared to GPT-3.5's 23-point difference). Consistent with existing literature, we see larger models outperforming their 7B counterparts and an overall performance decline across LLMs in the Indian context. A \emph{significant disparity persists between the Indic and non-Indic subsets,} with GPT-4-Turbo showing a 12-point difference between the two. Gemini 1.5 Flash exhibits an even more considerable gap, with a 32-point difference between the subsets. Additionally, in rationale prediction, LLMs tend to favor gold labels over their predictions when nudged for explanations. Interestingly, contrary to popular literature, we find the impact of chain-of-thought (CoT) prompting to be insignificant, reinforcing the challenging nature of \dataset.

\textbf{Contributions:} Through this work\footnote{Code and dataset available at \url{https://github.com/aflah02/QUENCH}}:
\begin{itemize}[noitemsep,nolistsep,topsep=0pt,leftmargin=*]
    \item We develop a novel open-domain quiz trivia dataset, \dataset, accompanied with rationales for each question.
    \item We benchmark \dataset\ on seven LLMs across $4$ evaluation metrics and $2$ prompting setups.
    \item We perform extensive analyses examining the influence of model size, prompting strategy, the role of indic vs. non-indic context, and highlight the most common prediction errors.
\end{itemize}

\section{Related Work}
Benchmarking and evaluating LLM is an active and evolving area of research \cite{10.1145/3641289, peng2024survey}. Benchmarks such as MMLU \cite{hendrycks2021measuring}, SuperGLUE \cite{NEURIPS2019_4496bf24}, HELM \cite{NEURIPS2023_dd83eada}, PromptBench \cite{zhu2023promptbench} and LMSys \cite{zheng2024lmsyschatm} provide a holistic suite of tasks to access the real-world adaptability of LLMs. 

Dedicated benchmarks have been proposed to access the mathematical \cite{lu-etal-2023-survey}, symbolic \cite{zhang2024llm}, commonsense/social \cite{10.1145/3615355,gandhi2023understanding}, and logical \cite{pan-etal-2023-logic,giadikiaroglou2024puzzle,sanyal-etal-2022-fairr} reasoning of LLMs. Datasets such as GSM-8k \cite{cobbe2021gsm8k}, JEEBench \cite{arora-etal-2023-llms}, MMLU \cite{hendrycks2021measuring}, MaScQA \cite{Zaki2024}, ScienceQA \cite{lu2022learn}, and LogiQA \cite{ijcai2020p501} have been designed to evaluate LLMs' knowledge across various subjects like mathematics, material science, and history. Most of these datasets utilize multiple-choice questions and focus on single themes per question. Concerns have been raised about the potential leakage of LLM pretraining data due to the public availability of these datasets \cite{xu2024benchmarking}. Additionally, there are datasets aimed at assessing commonsense reasoning \cite{zellers-etal-2019-hellaswag, Lourie2021UNICORNOR}, entity resolution \cite{10.1145/3474381}, and perceptiveness \cite{lin-etal-2022-truthfulqa}. 

Given that reasoning involves combining latent information of varying concepts \cite{wu-etal-2023-hence}, modality \cite{zhang2024multimodal, liu-etal-2022-things}, assessments in terms of knowledge-graph, and neurosymbolic \cite{olausson-etal-2023-linc} reasoning have also been proposed. However, these setups, too, tend to operate in an MCQ or cloze manner. Our work addresses the closed-knowledge gap by integrating world knowledge and commonsense reasoning into a quiz-based framework. This framework requires LLMs to infer masked entities within questions and generate rationales for their predictions. 

Meanwhile, for datasets specific to Indic cultures, such as \cite{seth-etal-2024-dosa, watts2024parikshalargescaleinvestigation}, there is a notable scarcity of challenging, open-ended, quiz-style benchmarks. Our work contributes to evaluating the performance of the LLMs under Indic and non-Indic setups. In the future, this can be extended to evaluate more fine-grained geographical and cultural setups. 

\section{QUENCH: Proposed Benchmark}
\label{sec:dataset}
\begin{figure}[!t]
    \centering
    \includegraphics[width=\columnwidth]{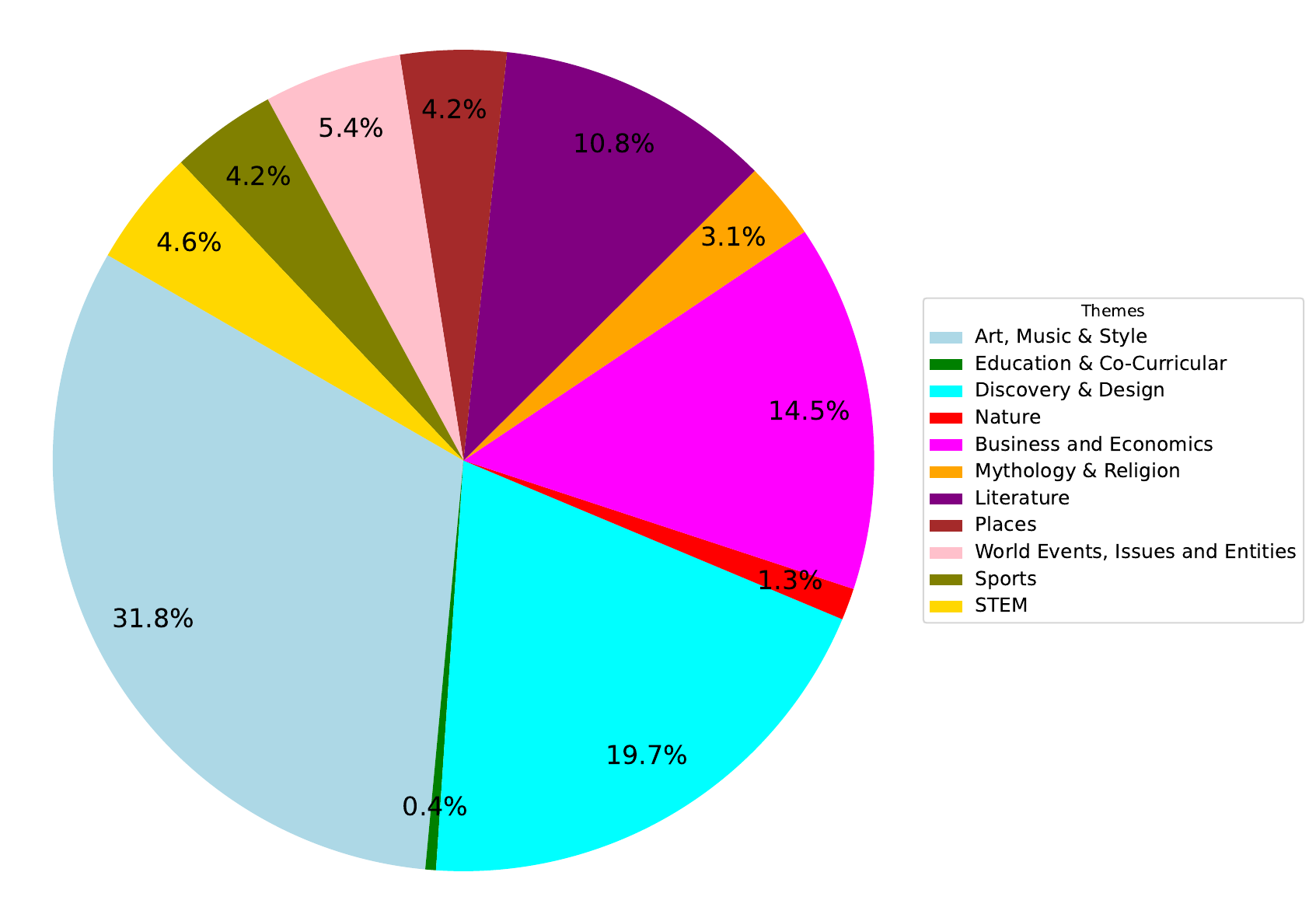}
    \caption{Themes and their distribution in \dataset.}
    \label{fig:theme-pie}
\end{figure}

\dataset\ is a collection of $400$ English questions from quiz competitions encompassing the $11$ themes in Figure \ref{fig:theme-pie}. Each question consists of a paragraph talking about some event related to one of the themes. In each question, some entities are masked with `X.' The aim is to connect the concepts in the question to predict `X.' The questions contain adequate cues to deduce the entities. The questions in the quiz already have the entities masked, and we have not modified these. However, we manually annotate the explanations/rationale to arrive at the answer. Some sample questions, answers, and rationales are provided in Table \ref{tab:quiz_examples}.

\paragraph{Data Sources.} Our primary source of questions is YouTube quizzing competitions videos \footnote{https://www.youtube.com/@KumarVarunOfficial}, with a tiny portion (7\%) from a website that publishes quizzing challenges\footnote{https://donquizote.wordpress.com/}. Upon exploring these two sources, we find that the videos provide more coherent reasoning and answers, better facilitating the annotation process. 

\begin{table}[!t]
\resizebox{\columnwidth}{!}{
\begin{tabular}{l|l|l|l|l|l}
\hline
\textbf{Subset} & \textbf{\# Q} & \textbf{\# E} & \textbf{Avg. QL} & \textbf{Avg. EL} & \textbf{Avg. RL} \\ \hline
\textbf{I}ndic & 70 & 80 & 77.03 & 1.85 & 39.48 \\
\textbf{N}on-\textbf{I}ndic & 330 & 379 & 84.82 & 1.96 & 40.52 \\
Over \textbf{A}ll & 400 & 459 & 83.46 & 1.94 & 40.33 \\ 
\hline
\end{tabular}
}
\caption{Dataset statistics of \dataset\ enlisting the number of questions (Q) and masked entities in the questions (E). We also report the average length of questions (QL), masked entities (EL), and the annotated rationale (RL).}
\label{tab:data_stat}
\end{table}

\begin{figure*}[t!]
    \centering
    \begin{subfigure}[c]{0.4\textwidth}
        \centering
        \includegraphics[height=2.6in]{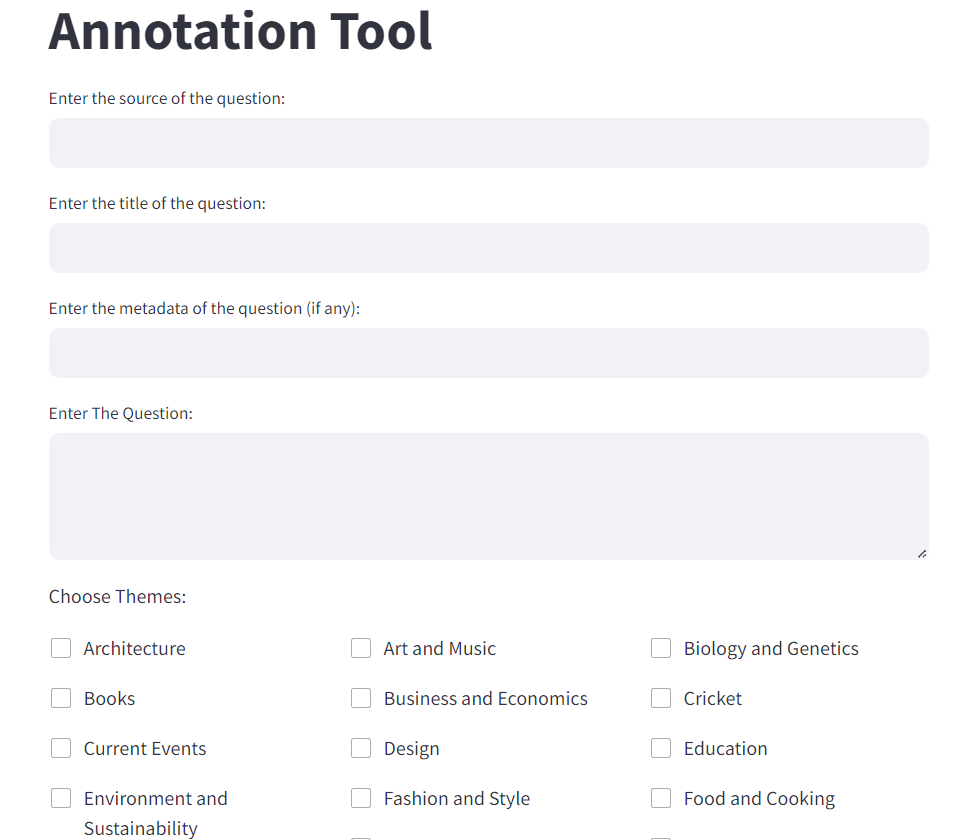}
        \caption{}
    \end{subfigure}
    ~ \hfill
    \begin{subfigure}[c]{0.4\textwidth}
        \centering
        \includegraphics[height=2.6in]{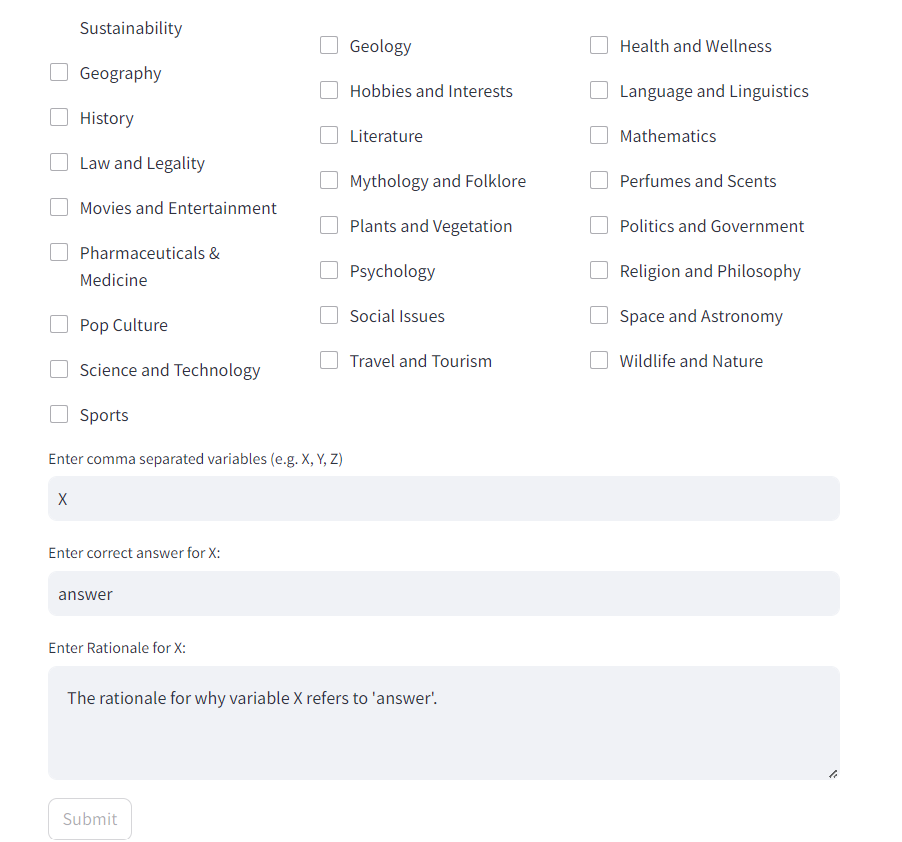}
        \caption{}
    \end{subfigure}
    \caption{Screenshots showing the upper (a) and lower (b) half of the custom annotation tool's landing page has $6 + 2 \cdot N$ sections to fill, where N is the number of masked entities in the question. The metadata section is optional.}
    \label{fig:annotation-tool}
\end{figure*}

\paragraph{Annotation Process.} The data annotation process is carried out by two male expert annotators (A1 and A2) aged 20-23. Both annotators possess
previous experience with cryptic hunts, quizzes, and similar activities. The annotators spend 8-10 minutes per question listening to the questions in the video. Working with the Text-Grab OCR tool\footnote{https://learn.microsoft.com/windows/powertoys/text-extractor}, the annotators extract the question text and answers for the masked entities. The annotators manually rectify any issues that arise. Based on the transcribed content of the video, the annotators paraphrase the rationale into coherent, point-wise sentences that outline how the correct entity can be deduced once the rationale is read. In cases where the rationale is not discussed in the video, the annotators are free to access the internet to obtain the explanations. The annotators then populate the following fields:
\begin{itemize}[noitemsep,nolistsep,topsep=0pt,leftmargin=2em]
    \item \textbf{Passage:} A passage with some entities/objects masked
    \item \textbf{List of Masked Entities:} The list of masked entities to be predicted. The questions are constructed to predict these entities.
    \item \textbf{List of Answers:} There is one answer for every mask entity.
    \item \textbf{List of Rationales:} The rationales behind each entity.
    \item \textbf{Themes:} A list of themes the passage fits in.
    \item \textbf{Source:} A URL to the question source.
\end{itemize}

\paragraph{Custom Annotation Tool.} The annotation process is carried out online without the need to scrape any videos. Once the transcription is obtained, the annotators compile the above information for each sample. We use a custom annotation tool with the help of Streamlit\footnote{https://streamlit.io/}. A screenshot of the annotation process is highlighted in Figure \ref{fig:annotation-tool}. The themes are multi-choice and subjective to the annotator's reasoning if not pre-defined. The variables section is constrained by the number of comma-separated variables inserted by the annotator. We also store the question's source so that anyone can proof-check these annotations if needed.

\begin{figure}[!t]
	\centering
	\includegraphics[scale=0.3]{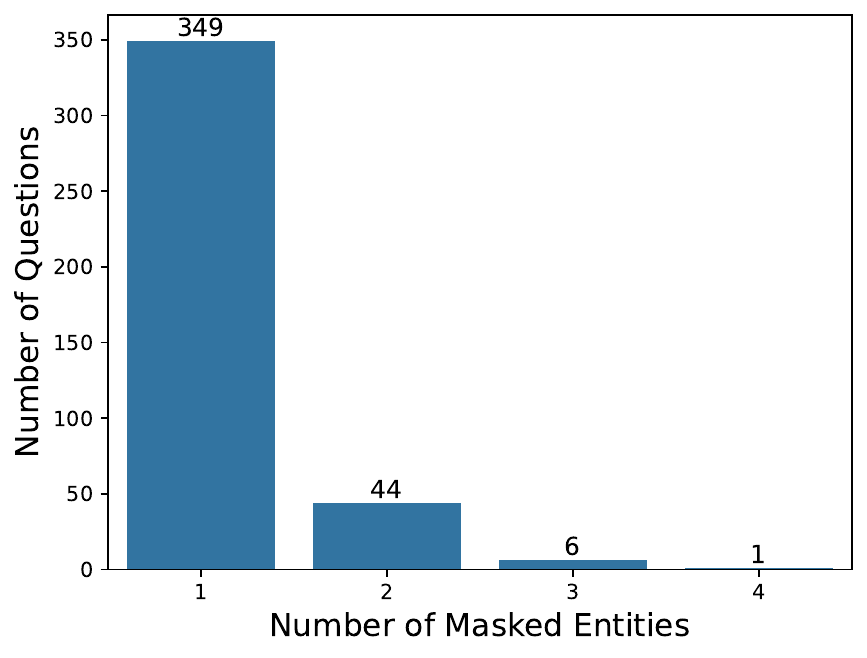}
	\caption{The number of masked entities in \dataset.}
	\label{fig:Number_of_Variables_in_Each_File}
\end{figure}

\paragraph{Curated Dataset.} The annotation process spanning 4 months is completed via a custom annotation tool as described below. We observe that a subset of questions exclusively pertains to Indian entities. Thus, we further tag each question as \textit{Indic} or \textit{Non-Indic}. The dataset statistics are outlined in Table \ref{tab:data_stat}. As a result of the annotations, the distribution of the number of masked entities across different questions is illustrated in Figure \ref{fig:Number_of_Variables_in_Each_File}.

\paragraph{Inter-annotator Agreement.} Since the rationale annotations are performed in a free-text manner, we carry out an inter-annotator evaluation to adjudge the quality of the rationale text. We randomly provide one annotator with $10$ samples compiled by the other. Both rank the free-text rationale on a 5-point Likert scale, with five being the highest quality of annotation. Based on this assessment, we obtain an inter-annotator agreement of $4.9$ from A1 to A2 and $4.85$ from A2 to A1.   

\paragraph{Unique Characteristics of \dataset.} \emph{Firstly, a critical aspect of our dataset is its objective yet open-domain nature.} While the answers to the quiz question are objective (one word/one phrase), the queries do not have fixed gold labels (for example, `Barack Hussein Obama,' `Barack Obama,' and `Obama' are all correct answers to the question `X was a civil rights attorney turned 44th President of USA'). Further, unlike the popular quiz show (Who Wants to Be Millionaire), we do not provide multiple-choice (MCQ) answers, which increases the difficulty of predicting the entities as the range of possibly correct entities is unrestricted. Simply removing the options in MCQ questions to produce a new dataset is not sufficient, as many of those questions depend on or refer to the options, such as, ``Which of the following is closest to X?". \emph{Secondly, the multi-hop reasoning and multiple co-reference resolution setup in \dataset\ spans intersectional themes within a question.} It provides a challenging environment to assess the world knowledge and entity recall capabilities of the LLMs.
\emph{Thirdly, it is also noteworthy that a subset of the quiz questions pertains to India-specific entities, allowing us to judge the indic-specific knowledge of the LLMs.} The questions mention sufficient Indian-specific concepts to nudge the deduction toward an India-specific answer without an explicit hint. \emph{Lastly, the dataset allows for a multifold assessment of LLMs covering both entity recall as well as rationale-building capabilities.} Overall, based on the above characteristics, it is evident \dataset\ provides a novel and challenging benchmark to access both world knowledge and recall capabilities of the LLM as well as establish the efficacy of the systems under open-domain setups beyond standardized NLP tasks. To ensure the quiz questions curated in \dataset\ are not already present in LLM pretraining datasets, we also conduct a data contamination check (Appendix \ref{app:data_leakage_analysis}). 

\section{Benchmarking Setup}
\label{sec:exp_setup}
This section outlines the models we employ for benchmarking \dataset\, along with the prompting and evaluation setups.

\paragraph{Benchmarked LLMs.} We experiment with various open-weight and closed-sourced instruction-tuned LLMs. Non-instruct LLMs are excluded from our assessment as they generate incoherent outputs. Similarly, formatting issues were registered from the Pythia family \cite{biderman2023pythia} models. Our model shortlisting was performed on 2xH100 and 2xA100 machines. The LLMs eventually shortlisted for benchmarking \dataset\ are furnished in Table \ref{tab:Models}. Amongst the closed-sourced models, we use the ones supported via APIs. Here, we employ \textbf{GPT-4-Turbo}, \textbf{GPT-3.5-Turbo}, and \textbf{Gemini-1.5-Flash}. For open-weighted models, we run inference via the free tier provided by Groq\footnote{https://console.groq.com/}. This setup allows fast inference at minimal infrastructural cost. Here we employ, \textbf{Meta-Llama-3-70B-Instruct}, \textbf{Meta-Llama-3-8B-Instruct}, \textbf{Mixtral-8x7B-Instruct-v0.1} and \textbf{Gemma-1.1-7B-Instruct}.

\begin{table}[!t]
\resizebox{\columnwidth}{!}{
\begin{tabular}{lcc}
\toprule
\textbf{Model} & \textbf{CW} & \textbf{Knowledge Cutoff} \\ \toprule
GPT-4-Turbo* (gpt-4-turbo-2024-04-09) & $128K$ & Dec'23 \\ 
GPT-3.5-Turbo* (gpt-3.5-turbo-0125) & $16k$ & Sep'21 \\
Gemini-1.5-Flash* & $1M$ & Nov'23 \\ 
Gemma-1.1-7B-Instruct$^\dagger$& $8K$ & Unknown \\ 
Mixtral-8x7B-Instruct-v0.1$^\dagger$ & $32k$ & Unknown \\ 
Meta-Llama-3-8B-Instruct$^\dagger$ & $8k$ & Mar'23 \\
Meta-Llama-3-70B-Instruct$^\dagger$ & $8k$           & Dec'23 \\ \bottomrule
\end{tabular}
}
\caption{Details of LLMs employed in this study. \textit{CW} captures the context window based on token length. Star(*) refers to closed-sourced LLMs and the dagger ($\dagger$) signifies open-weight LLMs.} 
\label{tab:Models}
\vspace{-5mm}
\end{table}

\paragraph{Prompting Predictions from LLMs.}
Given a quiz question and a list of missing entities, we prompt LLMs to predict the masked entities in the question. In the next stage of the pipeline, we use the predicted entities to prompt the model to provide a free text reason/rationale for why the expected entity correctly fits the context in the question. To separately examine the role of predicted entities in nudging the rationale, we also repeat the rationale prediction task with the gold-labeled entity to generate rationales. Here, we explore zero-shot prompting both with and without chain-of-thought (CoT) prompting \cite{wei2022chain}. For CoT prompting, we add the phrase ``Let's think step by step" similar to \citet{arora-etal-2023-llms} to the prompts outlined in Appendix \ref{app:prompt_llm}.

\input{overall_results_comb}

\paragraph{Evaluation Metrics.}
We employ a suit of natural language generation (NLG) based metrics for evaluation. Among the standard ones, we report BLEU \cite{Papineni02bleu}, ROUGE-L \cite{lin-2004-rouge}, and BERTScore \cite{bert-score} for semantic similarity. The standard automated metrics, however, will penalize the open-ended responses with variations like `U.S.A.' versus `United States' or `United States of America'. For fairer and more comprehensive analysis, we also explore LLM-driven evaluations. Recent works have shown that LLMs favor their own outputs \citep{LLM_Evaluators_Favor} during LLM-based evaluations. To address this, \citet{LLM_Jury} proposed using multiple LLMs as a jury. We, thus, employ the most robust models (GPT-4-Turbo, Mixtral-8x7B-Instruct-v0.1, and Meta-Llama-3-70B-Instruct) as our judges. Each judge individually scores the responses of every other benchmarked LLM, resulting in 21 judge-model combinations. In the case of entity prediction, we employ a binary scale to determine whether or not the entity is correct. For rationale prediction, we use a more granular 5-point Likert scale to capture semantics and handle nuances in lengthy texts. Since each question may contain multiple masked entities, we treat each entity as a distinct prediction. We aggregate the results across all entities for each question (\(\sum_{i=1}^{num\_q} \sum_{j=1}^{num\_ent(q_i)} score(q_i, ent_j)\)) where $num\_ent(q_i)$ counts how many masked entities are present in the question $q_i$ and finally scale the aggregated results to 100. The rationales are assessed under both predicted and gold entities. The evaluation prompts are outlined in Appendix \ref{app:eval_prompt}. 

\section{Results and Discussion}
\label{sec:results}
Comprehensive results for all metrics are provided in Appendix \ref{sec:results_appendix}, with a shorter aggregate provided in Table \ref{tab:overall-results-comb} for reference here. In each table, we refer to $\mathit{\mathbf{\Delta}}(N-I)$ as the difference in performance between the Non-Indic and Indic subsets.

\subsection{Entity Prediction}
In terms of standard metrics, we observe that GPT-4-Turbo is the best-performing model on \dataset, with Meta-Llama-3-70B-Instruct performing comparably well as the second-best model in terms of overall BERTScore. In Table \ref{tab:overall-results-comb}, GPT-4-Turbo produces the highest BERTScore of $97.3$ ($97.4$) when prompted without (with) CoT. Meta-Llama-3-70B-Instruct lags only by $\approx 1$ point with scores of $96.5$ ($96.5$) when prompted without (with) CoT. Similarly, in terms of LLM-Juries, we observe (Figure \ref{fig:combined-plots} (a)) that GPT-4-Turbo outperforms other models with a score of $86$ ($87$) when prompted without (with) CoT. Here again, the second-best models trail by $\approx 14$ points, with GPT-3.5-Turbo and Meta-Llama-3-70B-Instruct scoring $72$ ($73$) and $70$ ($72$) respectively when prompted without (with) CoT.

\input{figure-plots}

\subsection{Rationale Prediction}
Based on the entity predicted in the previous step, the LLMs are then prompted to explain how the predicted entities can be deduced/thought through. Interestingly, from Table \ref{tab:overall-results-comb}, we observe GPT-3.5-Turbo performs slightly better than the rest of the models, beating GPT-4-Turbo  and Meta-Llama-3-70B-Instruct by $\approx $ $0.2\%$ ($0.3\%$) and $0.2\%$ ($0.4\%$) BERTScores respectively, in terms of without (with) CoT. On the contrary, in terms of LLM-metrics (Figure \ref{fig:combined-plots} (b)), we observe that GPT-4-Turbo outshines other models in terms of producing coherent and correct rationals with scores as high as $90.41$ ($89.62$) for rationales generated without (with) CoT. The second-best systems trail by $\approx 10$ points, with GPT-3.5-Turbo and Meta-Llama-3-70B-Instruct scoring $80.32$ ($79.91$) and $80.15$ ($80.7$), respectively for without (with) CoT.

\paragraph{Note on standard metrics.} BERTScore is relatively more helpful than n-gram metrics like BLEU and ROUGE. \emph{Specifically, for rationale generation, syntactic metrics fail to convey meaningful information. This is because the generated rationales can vary significantly in structure, format, and length while conveying the same reasoning.} 

\subsection{Factors Influencing the Performance of Benchmarked LLMs}
We comment on the performance w.r.t standard and LLM-based metrics in terms of the training parameters, prompting strategy, the non-western context in the question, as well as the role of predicted/human entities in generating rationales.

\paragraph{Number of Parameters.} Across both standard and LLM-based metrics, there is a clear distinction in terms of the number of model parameters. Within the same family, Meta-Llama-3-70B-Instruct outperforms Meta-Llama-3-8B-Instruct on both entity and rationale predictions. However, this performance drop is not consistent even among models with a similar number of parameters. The variation among models with similar parameter sizes reiterates that not only just the number of parameters but also the pretraining strategy plays a role in the downstream reasoning ability of the LLMs.

\paragraph{Impact of Indic Subset.} All the benchmarking LLMs perform poorly at questions with an Indic context for both entity and rationale generation tasks. It is vital to reiterate that the only difference between indic and non-indic questions is the \textit{context}, as the dataset and predictions are curated in English. The performance variation is significantly noticeable across all metrics, as evident from Table \ref{tab:overall-results-comb}. In terms of GEval, the difference in performance is $\approx 21\%$ on average for predicting the correct entity (Figure \ref{fig:combined-plots} (c)) and $\approx 14.7\%$ for predicting the rationale (Figure \ref{fig:combined-plots} (d)) for the expected answer for non-Indic vs Indic entities. We hypothesize this is because the pretraining datasets for LLMs have a predominantly North American context \cite{zhou-etal-2022-richer}. \emph{It also means that cultural cues are critical while answering open-domain questions \cite{lee2024exploring}.}

\begin{figure*}[!t]
  \centering
  \begin{subfigure}[b]{0.32\textwidth}
    \centering
    \includegraphics[width=0.85\textwidth]{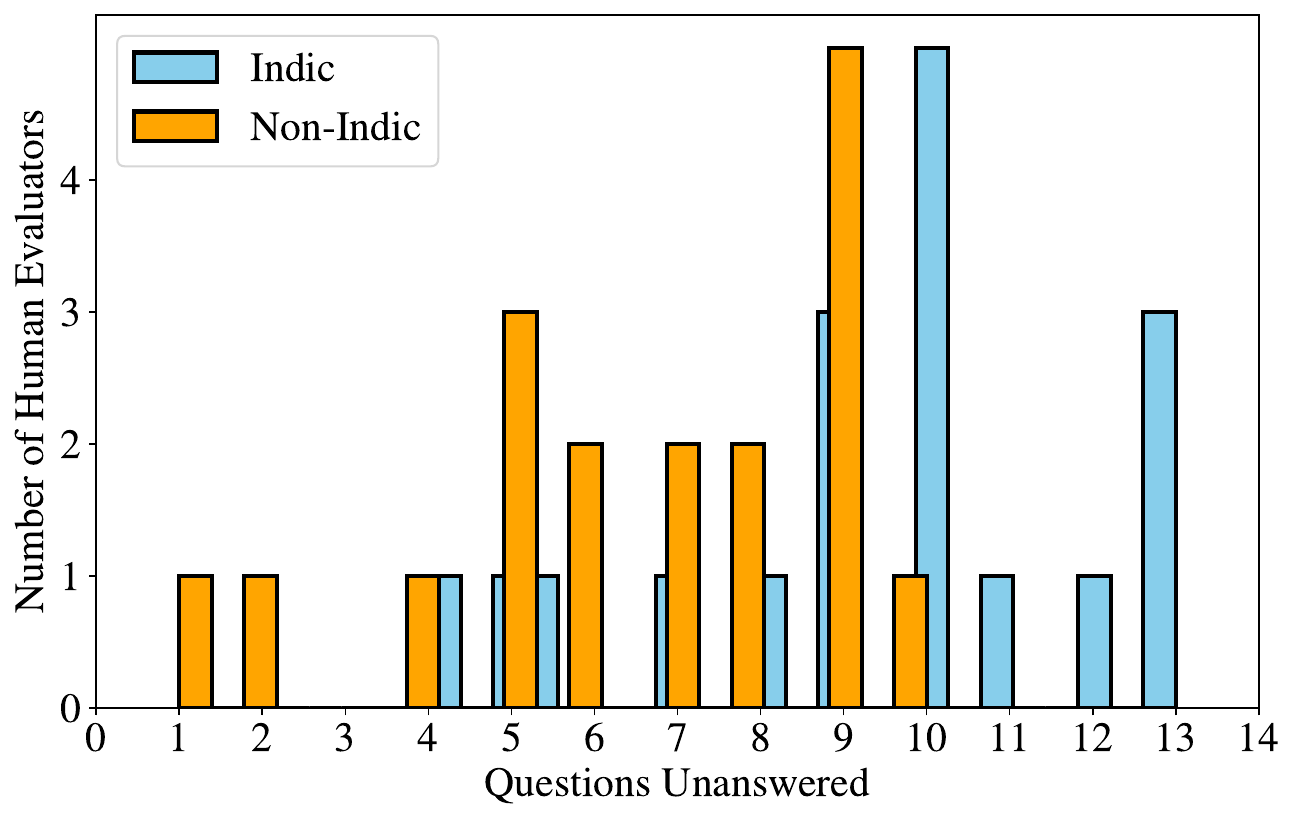}
    \caption{Unanswered Questions}
    \label{fig:unans_split}
  \end{subfigure}
  \begin{subfigure}[b]{0.32\textwidth}
    \centering
    \includegraphics[width=0.85\textwidth]{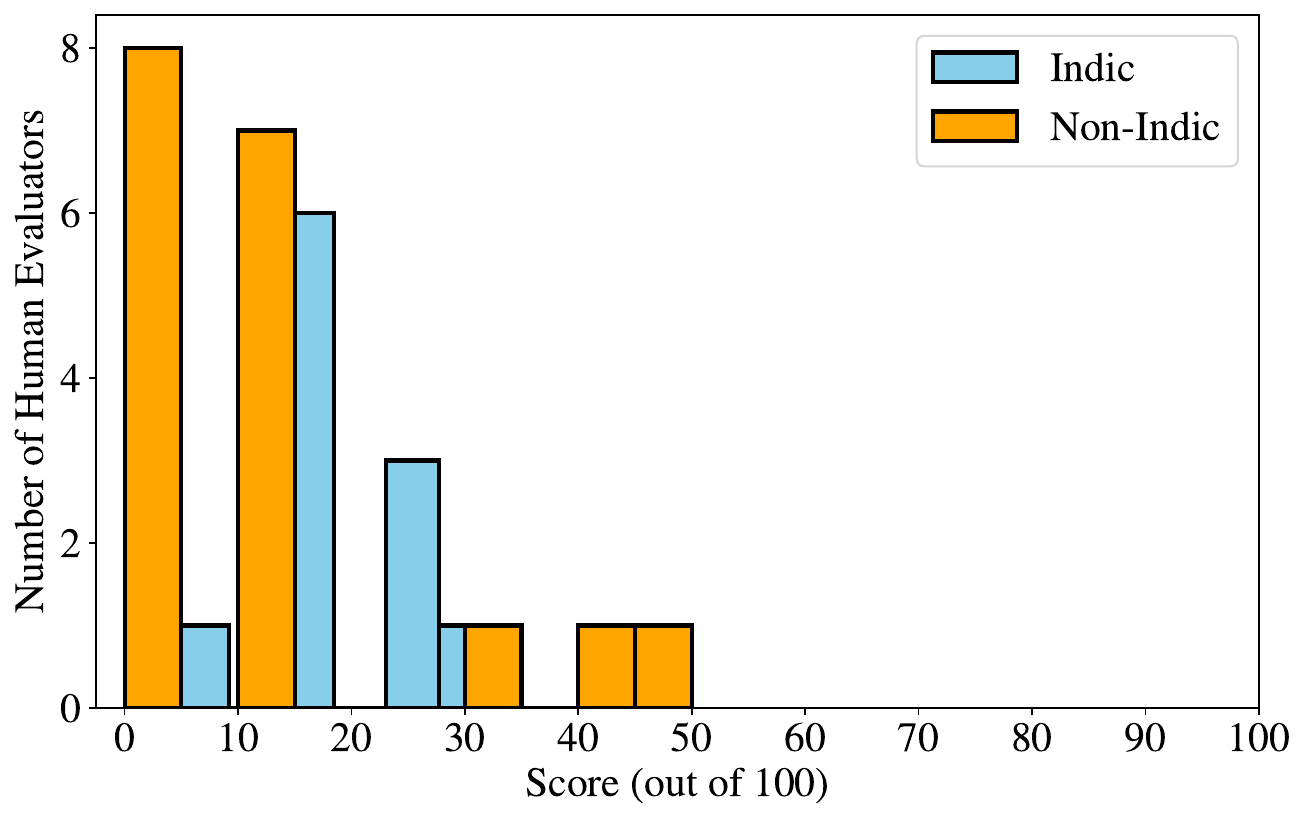}
    \caption{Entity prediction scores}
    \label{fig:split_ans_dist}
  \end{subfigure}
  \begin{subfigure}[b]{0.32\textwidth}
    \centering
    \includegraphics[width=0.85\textwidth]{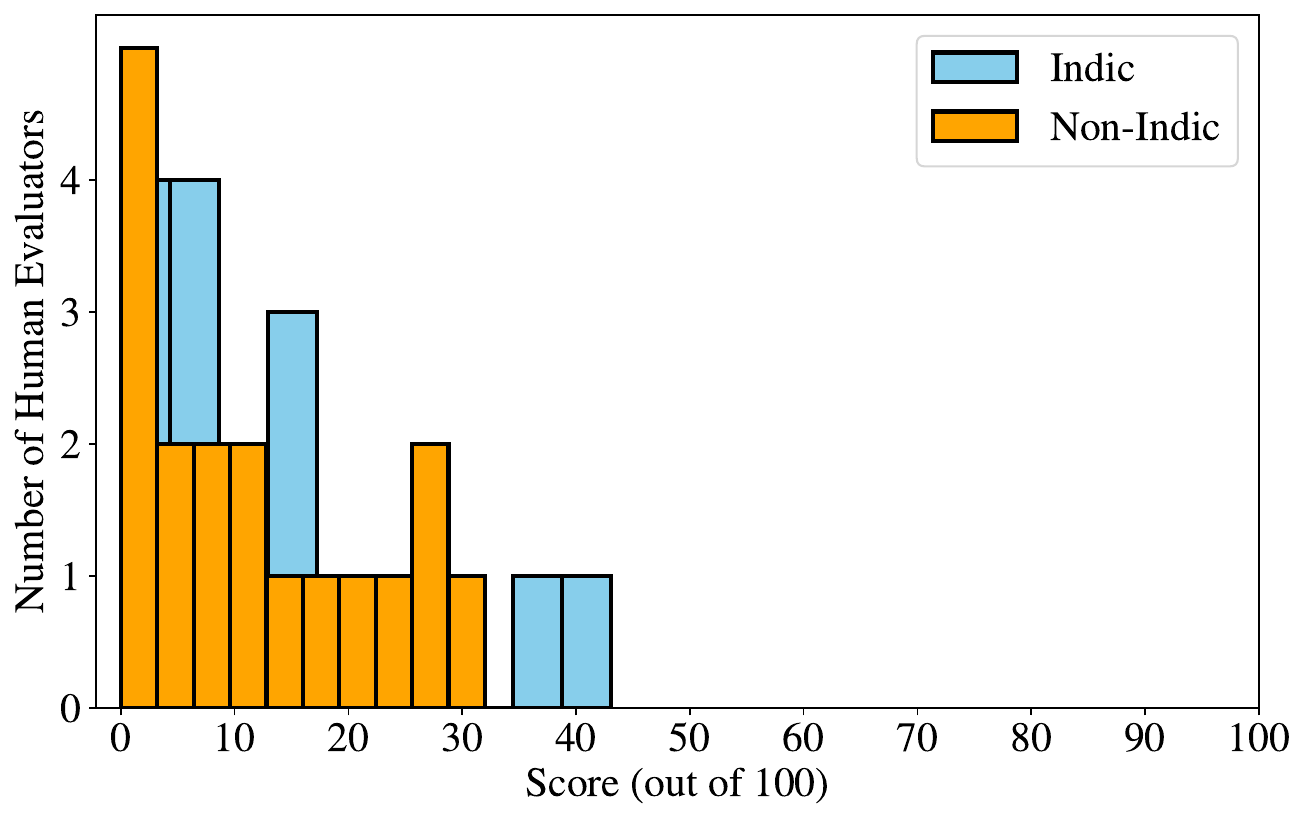}
    \caption{Rationale prediction scores}
    \label{fig:split_rationale_dist}
  \end{subfigure}
  \caption{Analysis of human benchmarking split across the indic and non-indic subsets capturing the distribution.}
  \label{fig:split_wise_analysis_ans_rationale}
\vspace{-3mm}
\end{figure*}

\paragraph{Rationale Predictions with Gold Labels.} In the case of generating a rationale for a given question, we alternatively provided the list of correct masked entities instead of the predicted ones. As expected, inserting gold labels in the prompts dramatically improves the quality of the rationale generated by the model. This difference is highest for Gemma-1.1-7b at $\approx 32\%$ (Tables \ref{tab:noncot-results-comb} and \ref{tab:cot-results-comb}). Similar behavior is observed even within Indic and non-Indic subsets. \emph{It shows that LLMs are capable of reasoning the path between the correct answer and question much better than generating the rationale behind their own predicted answers \cite{huang2024large}.}

\paragraph{CoT Prompting.} Contradictory to expected behavior, we observe via Figure \ref{fig:combined-plots} (a), (b) that the influence of CoT is inconsistent on \dataset. It may be a result of the challenging nature of the quiz-based task. LLMs need improvements in reasoning to make connections between multiple entities in the real world. \emph{In line with prior studies, we, too, observe that CoT optimization requires LLMs with parameters $>7B$ \cite{JMLR:v24:22-1144}.}

\section{Human Benchmarking}
To highlight the competitive nature of our benchmark, we also perform a human benchmarking.

\paragraph{Setup.} Given the resource-intensive nature of generating explainations, we randomly sample 20 questions for this assessment. We sample equal numbers from both subsets and across all themes. In total, we have 10 Non-Indic and 13 Indic questions, with some questions having more than one mask to predict. The participants answered both the missing entities and the rationale for the entities. The questions are distributed via a Google form (Appendix \ref{app:human_eval_benchmark}). The participants are given the option to respond with "NA" if they do not have an answer for an entity or rationale. This setup allows us to analyze refusal rates effectively. We recruit 18 participants for benchmarking. The participants consist of college students pursuing a range of degrees, from Bachelor's to Master's and PhD programs. 

\paragraph{Observations.} Based on the number of unanswered questions, Figure \ref{fig:split_wise_analysis_ans_rationale} (a) demonstrates the difficulty humans face with questions from \dataset. Moreover, both the task of predicting the correct answer and providing a rationale for it prove challenging, with the highest scores reaching only 30\%. Additionally, Figure \ref{fig:split_wise_analysis_ans_rationale} (b-c) reveals that participants generally find Indic questions more difficult to answer. Interestingly, there are more instances where Indic rationales receive higher scores, which might suggest that writing rationales for Indic questions is easier once an answer is known. However, due to the limited number of participants, we refrain from making any broad conclusions.

\section{Error Analysis}
\label{sec:errors}
Even the best-performing LLM (GPT-4-Turbo in our case) is prone to generative errors. In this section, we focus on the incorrect predictions arising from entity recognition and reasoning.

\paragraph{Incorrect Entity Recognition.} The model struggles with identifying the correct entities, even when using Chain-of-Thought (COT) reasoning. It becomes evident when, in 90.19\% of cases, the model receives the same jury score for both CoT and non-CoT generations.

\begin{table}[t]
    \resizebox{\columnwidth}{!}{
    \begin{tabular}{l|c|c}
        \toprule
        \multirow{2}{*}{\textbf{Error Types}} & \multicolumn{2}{c}{\textbf{Counts}} \\ \cline{2-3}
        & \textbf{w/o CoT} & \textbf{w/ CoT} \\\hline
        Unrelated to theme & 5 & 0 \\
        Unrelated but same theme & 10 & 0 \\
        Similar entities in same theme & 45 & 1 \\
        Wrong entity predicted but correct in rationale & 5 & 9 \\
        Correct Answer & 35 & 11 \\ \bottomrule
    \end{tabular}}
 \caption{Error analysis to classify each sample into one of the five error types (\texttt{a-e}, respectively).}
    \label{tab:error-types}
\end{table}

When prompted to identify entities within a specific domain, the model tends to favor well-known figures over the correct but less famous ones. For example, when asked about an Indian singer, it incorrectly identifies a widely recognized singer (e.g., ``Sonu Nigam'') instead of the correct and lesser well-known individual (e.g., ``Lucky Ali''). Consequently, the model has a tendency to link entities to more common or prominent organizations within a field rather than accurately identifying the rare entity. For instance, in a scenario where the correct answer is ``Amrutanjan'' (a patent medicine business), the model incorrectly identifies ``All India Radio,'' possibly due to the mention of a journalist in the question. The model can provide misinformed entities, revealing significant gaps in its knowledge and ability to differentiate between similarly categorized entities. For example, it incorrectly identifies ``Yogi Adityanath'' (the Chief Minister of Uttar Pradesh) as the Chief Minister of Odisha instead of ``Naveen Patnaik.''

Despite the shortcomings in specific entity recognition, the model shows a surprising proficiency in identifying high-level relationships. Suppose the answer involves identifying a cricketer. In that case, the model generally generates a name within the correct category (i.e., a cricketer), even if it does not pinpoint the exact individual. It suggests that while the model has a reasonable grasp of broad categories, roles, and domain leaders within that role, it may struggle with precise identification within those categories. Some other examples are predicting ``Waheeda Rehman'' in place of ``Bhanu Athaiya,'' who are both famous Indian actresses.

\paragraph{Incorrect Rationale Generation:} LLMs can also struggle to generate specific rationales. We conduct a 2-dimensional analysis of GPT-4 over 100 randomly sampled predictions and record:
\begin{enumerate}[noitemsep,nolistsep,topsep=0pt,leftmargin=2em]
    \item The type/category of error in the generations outlined in (Table \ref{tab:error-types}).
    \item If the model is able to generate the correct rationale even if the entity predicted is incorrect as outlined in (Table \ref{tab:rationale-errors}).
\end{enumerate}

\begin{table}[t]
\centering
\resizebox{\columnwidth}{!}{
    \begin{tabular}{lcc}
        \toprule
        & \multicolumn{2}{c}{\textbf{Correct Rationale for incorrect prediction?}} \\ \cline{2-3}
        & \textbf{Yes} & \textbf{No} \\ \hline
        \textbf{Counts (w/o CoT)} & 17 & 83 \\ 
        \textbf{Counts (w/ CoT)} & 8 & 2 \\ \bottomrule
    \end{tabular}}
    \caption{Assessing when LLMs produce a correct rationale despite making an incorrect entity prediction.}
    \label{tab:rationale-errors}
\end{table}

Table \ref{tab:error-types} shows the statistics for the categorical error analysis. We highlight that the CoT sample count does not total 100 because we only mark samples where the reasoning differed significantly between the two experiments (without and with CoT). We see that most of the errors are of type \texttt{c}, where the theme is correct, but the LLM gets confused between two very similar entities from the same domain (possibly due to differences in entity popularity/richness of embeddings or their closeness in terms of semantic overlap). 

Through manual assessment, we also notice that CoT may not be effective in the entity prediction part. When employed for rationale generation, it can rectify the mistakes in initial entity predictions. Table \ref{tab:rationale-errors} shows that for $\approx 20\%$ of the subset of samples, the model is able to predict the rationale correctly without needing first to predict the correct entity, and for $80\%$ of the anomalous cases in CoT, we see that the rationales are correct. These observations show that LLMs are incredibly sensitive to the way prompts are framed, and predicting entities in a subjective fashion is a relatively more complex challenge than in an MCQ setting. 

\section{Conclusion}
We devise a novel benchmark, \dataset\, of about 400 quizzing questions in English from a diverse set of themes. \dataset\ helps evaluate the deductive reasoning capabilities of LLMs. Interestingly, by accessing CoT and without CoT prompting techniques, we recorded the non-consequent differences between the two setups for our dataset. We also find that LLMs are much better at answering questions that have a general/non-indic context. Overall, we observe \dataset\ to be a challenging benchmarking necessitating future research in the area of open-domain deductive reasoning. In the future, we would like to extend our assessment under multilingual settings and benchmark other reasoning techniques such as tree-of-thoughts, question decomposition, and self-consistent CoT.

\section*{Limitations}
Despite our best efforts, we could not evaluate a broader range of models and prompting techniques due to resource constraints. This study examines the behavior of entity and rationale predictions in a 2-step fashion, and the real impact of CoT may come into play if both tasks are prompted in a single prompt with the liberty first to rationalize and then predict the entities. However, parsing such responses will be complex as a strict format may not be followed, and it may require increased human efforts for evaluation. While a jury can lead us to a better assessment, each jury-LLM incurs a cost in terms of hardware resources and API. 

\section*{Ethical Considerations}
Human annotators play a crucial role in the development of our dataset. We ensure that all annotators are fairly compensated for their work and provided with clear instructions to minimize subjectivity and bias in their annotations. Further, the annotators were offered sufficient time to annotate so as not to burden them. Annotators were also given the option to decline participation without any repercussions. We maintained a respectful and supportive work environment throughout the annotation process. Secondly, we source all our data from public platforms and test both closed-source and open-weight models, which allow for a fair benchmarking of LLMs. Lastly, we also make minimal use of LLMs for re-writing and grammatical corrections and, in some cases, Copilot for code completion during experiments.

\section*{Acknowledgments}
The authors acknowledge the support of the Infosys Foundation through the Center for AI (CAI) at IIIT Delhi.


\bibliography{custom}
\clearpage
\newpage
\appendix
\section{Data Leakage Analysis}
\label{app:data_leakage_analysis}

Due to the potential for benchmark leakage into pretraining corpora, we investigate several popular data sources to ensure our benchmark remains unaffected. 

\subsection*{Phase 1: Checking for presence of data sources in pretraining corpora}
Specifically, we examine \url{youtube.com} (YT) and \url{donquizote.wordpress.com} (DQ), which are primary sources using \citet{elazar2024whatsbigdata}, to determine if data from these sources has been used in pretraining. We check for contamination in the C4 \cite{c4-t5}, mC4-en \cite{mc4en}, OSCAR \cite{oscar}, RedPajama \cite{together2023redpajama}, LAION-2B-en \cite{schuhmann2022laion} and Dolma \cite{soldaini-etal-2024-dolma} datasets.

We observe a negligible presence of both data sources across all public pretraining corpora, with a maximum of around 22K tokens, and most instances below 1K tokens. Notably, older datasets like OSCAR and C4 show a slightly higher presence of DQ, with 23K tokens (~0.0000046\% of total tokens) and 9K tokens (~0.00000032\% of total tokens), respectively. In contrast, newer datasets such as Dolma contain only 40 tokens (~0.00000091\% of total tokens). Other datasets show no trace of the site.

Similarly, YT follows the same pattern. WIMBD's prefix search also picks up other URLs like \url{youtube.comactivate.org}, which share the "youtube.com" prefix, but their presence is minimal (<0.00000001\%). The highest YT presence is found in mC4-en with 22.3K tokens (~0.00000081\% of total tokens), indicating that both sources have an insignificant presence in the pretraining corpora.

\subsection*{Phase 2: Checking for exact match count in pretraining corpora}
We utilize the Infinigram API \citep{Liu2024InfiniGram} to search for exact question matches within the following pretraining corpora: Dolma \cite{soldaini-etal-2024-dolma}, RedPajama \cite{together2023redpajama}, Pile \cite{gao2020pile800gbdatasetdiverse}, and C4 \cite{c4-t5}. Our analysis shows zero contamination across all corpora, confirming that our dataset has not been leaked and any model performance on it is unrelated to memorization.

\section{Prediction Prompts}
\label{app:prompt_llm}
\begin{itemize}[noitemsep,nolistsep,topsep=0pt,leftmargin=2em]
    \item \textbf{For predicting entities:} \textit{Consider yourself a participant in a quiz show where I am the quizmaster. I will ask you a question that can be from any general theme. You need to provide me with the correct answer. The question can have multiple variables to answer, and you need to provide me with the answer for variable \{\}. Hence, use the following format strictly in your response: `The answer is <X answer>.' You lose points if you fail to follow the format.}
    \item \textbf{For generating rationale using prediction:} \textit{Consider yourself a participant in a quiz show where I am the quizmaster. I will ask you a question that can be from any general theme. The prediction for variable \{\} is \{\}. Provide me with the rationale followed for your answer. Use the following format in your response: `The rationale is <rationale>'. You lose points if you fail to follow the format."}
    
    \item \textbf{For generating rationale using gold labels:} \textit{Consider yourself a participant in a quiz show where I am the quizmaster. I will ask you a question that can be from any general theme. You need to provide me with the correct answer. The question can have multiple variables to answer, and you need to provide me with the answer for variable \{\}. Hence, use the following format strictly in your response: `The answer is <X answer>.' You lose points if you fail to follow the format.}
\end{itemize}
Note that the phrase ``You lose points if you fail to follow the format'' is only added to induce a quizzing setup. Since ours is a zero-shot setup, there is no explicit punishment/loss sent as feedback to the model if the answer is not correctly predicted. Further, the models are evaluated independently and not competing against each other.

\section{Evaluation Prompts}
\label{app:eval_prompt}
\begin{itemize}[noitemsep,nolistsep,topsep=0pt,leftmargin=2em]
    \item \textbf{Entity Evaluation:} \textit{You are the host of a quiz show where you ask complex and tricky questions to the contestants. Now, once you ask one such question, the contestant gives an answer that might not be the exact answer but is still correct. For instance, the answer provided to you might be 'U.S.A' while the actual answer is 'United States of America' or 'United States' or 'America', etc. Use your wise judgment to decide, based on the question given, whether the answer is correct or not. YOU ARE THE JUDGE AND YOUR WORD IS FINAL. Be fair and just in your judgment. Always respond with 'correct' or 'incorrect' based on the answer provided by the contestant. You will be provided the question, the true answer, and the answer provided by the contestant, and you need to decide whether the answer is correct or not.}
    \#\# Question
    
<question>

\#\# True Answer

<true\_answer>

\#\# Answer Given by contestant

<answer\_given\_by\_contestant>

\#\# Your Judgement (correct/incorrect)

    \item \textbf{Rationale Evaluation:} 

\textit{You are the host of a quiz show where you ask complex and tricky questions to the contestants. Now, once you ask one such question, the contestant gives an answer as well as the rationale behind that answer. You need to decide whether the rationale provided is correct or not by comparing it with the true rationale. Use your wise judgment to decide based on the question given whether the rationale is correct or not. YOU ARE THE JUDGE AND YOUR WORD IS FINAL. Be fair and just in your judgment. Provide a score between 1 to 5 based on the rationale provided by the contestant. 1 being the least and 5 being the highest score.
}
\#\# Question

<question>

\#\# True Rationale

<true\_answer>

\#\# Rationale Given by contestant

<rationale\_given\_by\_contestant>

\#\# Your Judgement (Score between 1 to 5 do not provide any other score or text)

\end{itemize}

\section{Evaluation Results}
\label{sec:results_appendix}
Tables \ref{tab:cot-results-comb} and \ref{tab:noncot-results-comb} capture the results with and without CoT prompting on varying subsets of the dataset and record the evaluation on all metrics BLEU, ROUGE, BERTScore and LLM-metrics.

\textbf{BLEU and ROUGE.} From Tables \ref{tab:noncot-results-comb} and \ref{tab:cot-results-comb},
with a BLEU of $65.7$ ($66.2$), GPT-4-Turbo leads among all models when prompted without (with) CoT. In comparison, Meta-Llama-3-70B-Instruct trails by $\approx 4$ points with a BLEU of $61.2$ ($61.5$) when prompted without (with) CoT. Meanwhile, in terms of ROUGE-L (Tables \ref{tab:noncot-results-comb} and \ref{tab:cot-results-comb}), we observe similar patterns. GPT-4-Turbo leads among all models when prompted without (with) CoT with a score of $89.8$ ($89.9$), and Meta-Llama-3-70B-Instruct trails behind with $83.3$ ($83.6$) when prompted without (with) CoT.  On the other hand, GPT-3.5-Turbo leads with $22.5$ ($22.7$) BLEU and  $29.1$ ($29.7$) ROUGE-L when prompted without (with) CoT. Meta-Llama-3-70B-Instruct trails by $1$ points with a BLEU of $21.4$ ($21.3$) and a ROUGE-L of $28.7$ ($28.7$)
when prompted without (with) CoT. 

\textbf{Tradeoff between Closed Source v/s Open Weight Models.} Across all metrics and experiments, GPT-4-Turbo consistently outperformed other models. While its performance is closely followed by both GPT-3.5-Turbo and the open-weighted Meta-Llama-3-70B-Instruct, the tradeoff between using open and closed sources is apparent. Speaking broadly about the LLM-evaluated metrics, if the only parameter to optimize is performance (in lieu of cost or open access), then GPT-4-Turbo should be the preferred model for world knowledge-related tasks. Meanwhile, Meta-Llama-3-70B-Instruct can efficiently serve as a substitute for the closed-sourced API models. It registers a slight drop in performance against GPT-4-Turbo and clearly outperforms Gemini-1.5-Flash, while performance is very comparable to GPT-3.5-Turbo. However, the $70B$ model comes with resource constraints. In terms of the LLM-evaluated metric, Mixtral-8x7B-Instruct-v0.1 is an excellent alternative to Meta-Llama-3-70B-Instruct in case of limited hardware, albeit with a significant performance drop. 

\begin{figure*}[t!]
\centering
    \begin{subfigure}[c]{0.4\textwidth}
        \centering
        \includegraphics[height=2.6in]{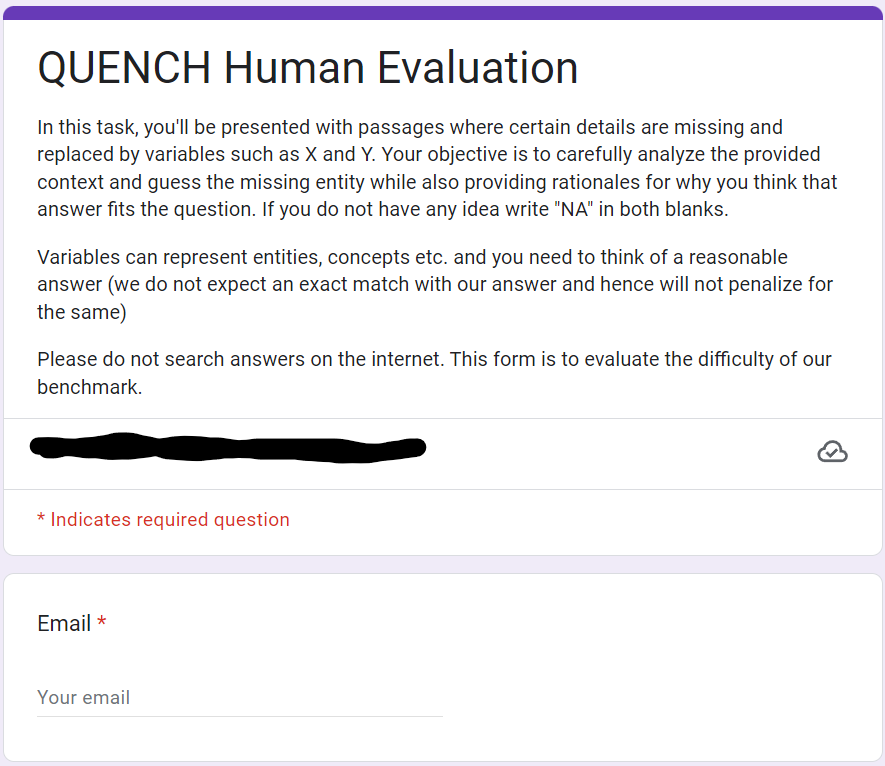}
        \caption{}
    \end{subfigure}
    ~ \hspace{5em}
    \begin{subfigure}[c]{0.4\textwidth}
        \centering
        \includegraphics[height=2.6in]{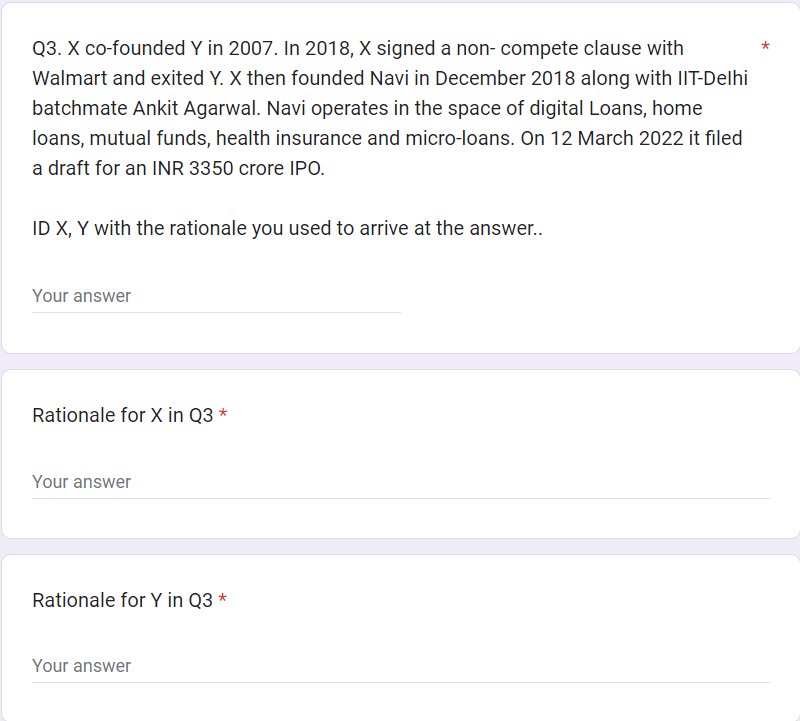}
        \caption{}
    \end{subfigure}
    \caption{Screenshots showing the instructions (a) and one of the questions with multiple rationales (b) for the human evaluation of \dataset.}
    \label{fig:humaneval_form}
\end{figure*}

\input{result-benchmark}

\section{Human Benchmarking}
\label{app:human_eval_benchmark}
Figure \ref{fig:humaneval_form} provides an overview of the Google form employed for human benchmarking of \dataset.
\end{document}

%% file: overall_results_comb.tex
\begin{table*}
\tiny
\resizebox{\textwidth}{!}{
\begin{tabular}{llrrrrrrrr}
\toprule
\multirow{3}{*}{\textbf{Model}} & \multirow{3}{*}{\textbf{Category}} & \multicolumn{4}{c}{\textbf{Without Chain-of-Thought}} & \multicolumn{4}{c}{\textbf{With Chain-of-Thought}} \\ \cline{3-10} 
 &  & \multicolumn{2}{c}{\textbf{Answer}} & \multicolumn{2}{c}{\textbf{Rationales with Predicted Answer}} & \multicolumn{2}{c}{\textbf{Answer}} & \multicolumn{2}{c}{\textbf{Rationales with Predicted Answer}} \\ \cline{3-10}
 &  & \multicolumn{1}{c}{BS} & \multicolumn{1}{c}{GEval} & \multicolumn{1}{c}{BS} & \multicolumn{1}{r}{GEval} & \multicolumn{1}{c}{BS} & \multicolumn{1}{c}{GEval} & \multicolumn{1}{c}{BS} & \multicolumn{1}{r}{GEval} \\ \midrule
\multirow{4}{*}{Gemini 1.5 Flash} & Indic & 91.7 & \multicolumn{1}{r}{40} & 86.8 & 56.2 & 91.3 & \multicolumn{1}{r}{38} & 86.7 & 56.4 \\
 & Non-Indic & 93.6 & \multicolumn{1}{r}{71} & 86.9 & 76 & 93.7 & \multicolumn{1}{r}{70} & 87.1 & 75.4 \\
 & All & 93.3 & \multicolumn{1}{r}{66} & 86.9 & 72.6 & 93.3 & \multicolumn{1}{r}{64} & 87.1 & 72.2 \\ \cdashline{2-10}
 & $\pm \mathit{\mathbf{\Delta}}(N-I)$ & $+1.9$ & \multicolumn{1}{r}{$+31.0$} & $+0.1$ & $+19.8$ & $+2.4$ & \multicolumn{1}{r}{$+32.0$} & $+0.4$ & $+19.0$ \\ \midrule
\multirow{4}{*}{Gemma-1.1-7B-it} & Indic & 89.2 & \multicolumn{1}{r}{14} & 86 & 38.6 & 87.5 & \multicolumn{1}{r}{20} & 86.2 & 43.6 \\
 & Non-Indic & 90.8 & \multicolumn{1}{r}{41} & 85.9 & 58.2 & 89.3 & \multicolumn{1}{r}{43} & 86.1 & 59.6 \\
 & All & 90.6 & \multicolumn{1}{r}{36} & 86 & 54.8 & 89 & \multicolumn{1}{r}{39} & 86.1 & 56.8 \\ \cdashline{2-10}
 & $\pm \mathit{\mathbf{\Delta}}(N-I)$ & $+1.6$ & \multicolumn{1}{r}{$+27.0$} & $-0.1$ & $+19.6$ & $+1.8$ & \multicolumn{1}{r}{$+23.0$} & $-0.1$ & $+16.0$ \\ \midrule
\multirow{4}{*}{GPT-3.5-Turbo} & Indic & 95 & \multicolumn{1}{r}{53} & 87.4 & 67.2 & 94.6 & \multicolumn{1}{r}{54} & 87.6 & 66.2 \\
 & Non-Indic & 95.9 & \multicolumn{1}{r}{76} & 87.6 & 83 & 96 & \multicolumn{1}{r}{77} & 87.6 & 82.6 \\
 & All & 95.8 & \multicolumn{1}{r}{72} & 87.5 & 80.2 & 95.8 & \multicolumn{1}{r}{73} & 87.6 & 79.8 \\ \cdashline{2-10}
 & $\pm \mathit{\mathbf{\Delta}}(N-I)$ & $+0.9$ & \multicolumn{1}{r}{$+23.0$} & $+0.2$ & $+15.8$ & $+1.4$ & \multicolumn{1}{r}{$+23.0$} & $0.0$ & $+16.4$ \\ \midrule
\multirow{4}{*}{GPT-4-Turbo} & Indic & 96.8 & \multicolumn{1}{r}{78} & 87.5 & 86.8 & 97 & \multicolumn{1}{r}{77} & 87.5 & 82.8 \\
 & Non-Indic & 97.4 & \multicolumn{1}{r}{88} & 87.3 & 91 & 97.5 & \multicolumn{1}{r}{89} & 87.3 & 91 \\
 & All & 97.3 & \multicolumn{1}{r}{86} & 87.3 & 90.4 & 97.4 & \multicolumn{1}{r}{87} & 87.3 & 89.6 \\ \cdashline{2-10}
 & $\pm \mathit{\mathbf{\Delta}}(N-I)$ & $+0.6$ & \multicolumn{1}{r}{$+10.0$} & $-0.2$ & $+4.2$ & $+0.5$ & \multicolumn{1}{r}{$+12.0$} & $-0.2$ & $+8.2$ \\ \midrule
\multirow{4}{*}{Meta-Llama-3-8B-Instruct} & Indic & 94 & \multicolumn{1}{r}{29} & 86.3 & 48.8 & 93.3 & \multicolumn{1}{r}{25} & 86.5 & 46.6 \\
 & Non-Indic & 95.3 & \multicolumn{1}{r}{46} & 86.4 & 60.8 & 95.4 & \multicolumn{1}{r}{47} & 86.4 & 62.6 \\
 & All & 95.1 & \multicolumn{1}{r}{43} & 86.3 & 58.6 & 95 & \multicolumn{1}{r}{43} & 86.4 & 59.8 \\ \cdashline{2-10}
 & $\pm \mathit{\mathbf{\Delta}}(N-I)$ & $+1.3$ & \multicolumn{1}{r}{$+17.0$} & $+0.1$ & $+12.0$ & $+2.1$ & \multicolumn{1}{r}{$+22.0$} & $-0.1$ & $+16.0$ \\ \midrule
\multirow{4}{*}{Meta-Llama-3-70B-Instruct} & Indic & 95.4 & \multicolumn{1}{r}{58} & 87 & 70.8 & 95.7 & \multicolumn{1}{r}{62} & 87 & 71.8 \\
 & Non-Indic & 96.7 & \multicolumn{1}{r}{73} & 87.3 & 82 & 96.7 & \multicolumn{1}{r}{74} & 87.3 & 82.4 \\
 & All & 96.5 & \multicolumn{1}{r}{70} & 87.3 & 80 & 96.5 & \multicolumn{1}{r}{72} & 87.2 & 80.6 \\ \cdashline{2-10}
 & $\pm \mathit{\mathbf{\Delta}}(N-I)$ & $+1.3$ & \multicolumn{1}{r}{$+15.0$} & $+0.3$ & $+11.2$ & $+1.0$ & \multicolumn{1}{r}{$+12.0$} & $+0.3$ & $+10.6$ \\ \midrule
\multirow{4}{*}{Mixtral-8x7B-Instruct-v0.1} & Indic & 86.1 & \multicolumn{1}{r}{43} & 87 & 61.4 & 86 & \multicolumn{1}{r}{42} & 86.9 & 59.2 \\
 & Non-Indic & 88 & \multicolumn{1}{r}{68} & 87 & 79.4 & 88.4 & \multicolumn{1}{r}{71} & 87.1 & 78.4 \\
 & All & 87.7 & \multicolumn{1}{r}{64} & 87 & 76.2 & 88 & \multicolumn{1}{r}{66} & 87 & 75 \\ \cdashline{2-10}
 & $\pm \mathit{\mathbf{\Delta}}(N-I)$ & $+1.9$ & \multicolumn{1}{r}{$+25.0$} & $0.0$ & $+18.0$ & $+2.4$ & \multicolumn{1}{r}{$+29.0$} & $+0.2$ & $+19.2$ \\
 \bottomrule

\end{tabular}
}
\caption{LLM performances on \dataset\ with and without Chain-of-Thought prompting. Here, $\mathit{\mathbf{\Delta}}(N-I)$ is the difference in performance between the Non-Indic and Indic subset. BS: BERTScore; GEval: Jury Evaluation.}
\label{tab:overall-results-comb}
\vspace{-3mm}
\end{table*}

%% file: figure-plots.tex
\begin{figure*}[t!]
    \centering
    \begin{subfigure}[t]{0.45\textwidth}
        \centering
        \includegraphics[width=\textwidth]{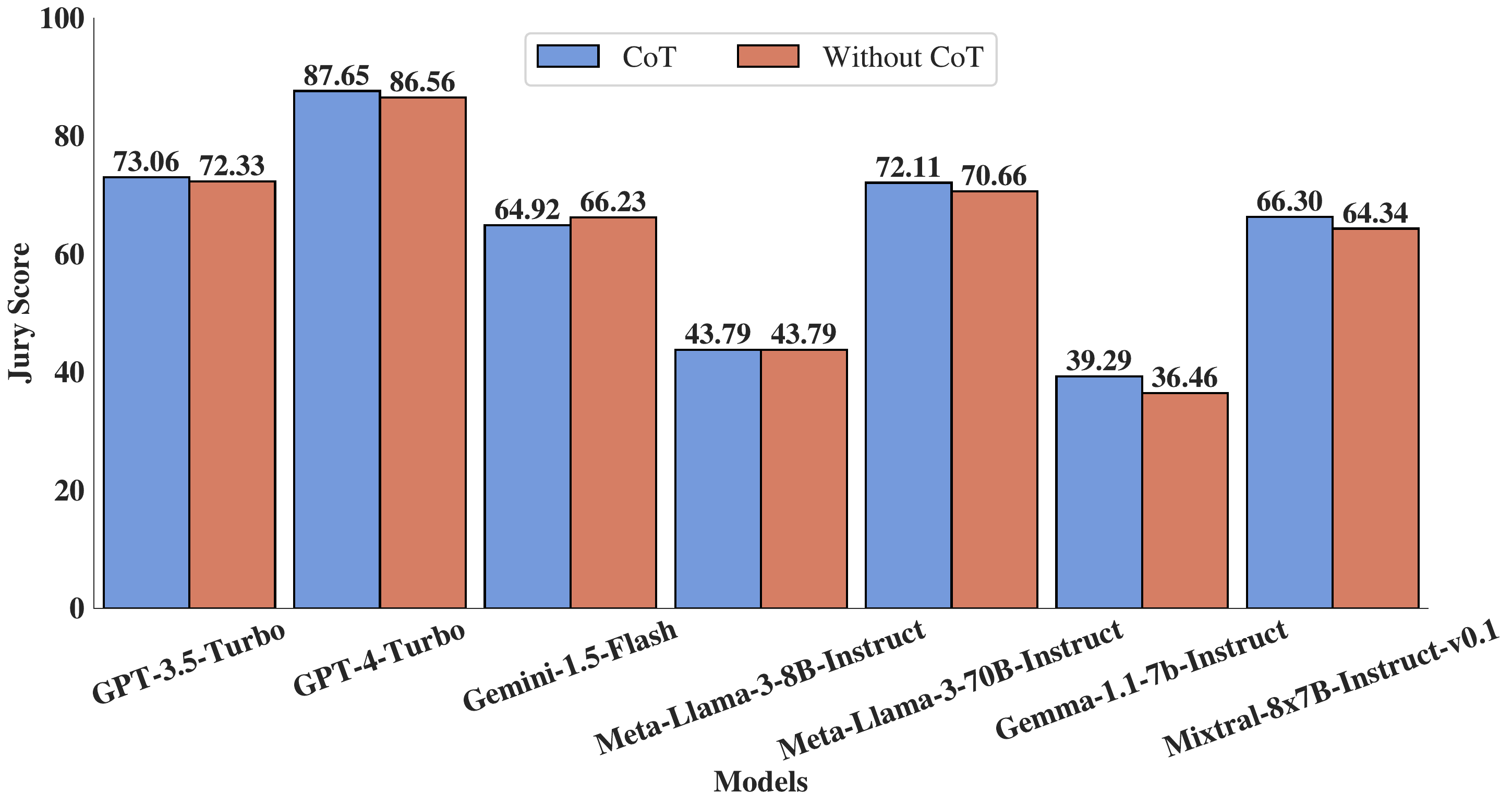}
        \label{fig:overall-correct}
        \caption{Predicted answer.}
    \end{subfigure}
    ~ 
    \begin{subfigure}[t]{0.45\textwidth}
        \centering
        \includegraphics[width=\textwidth]{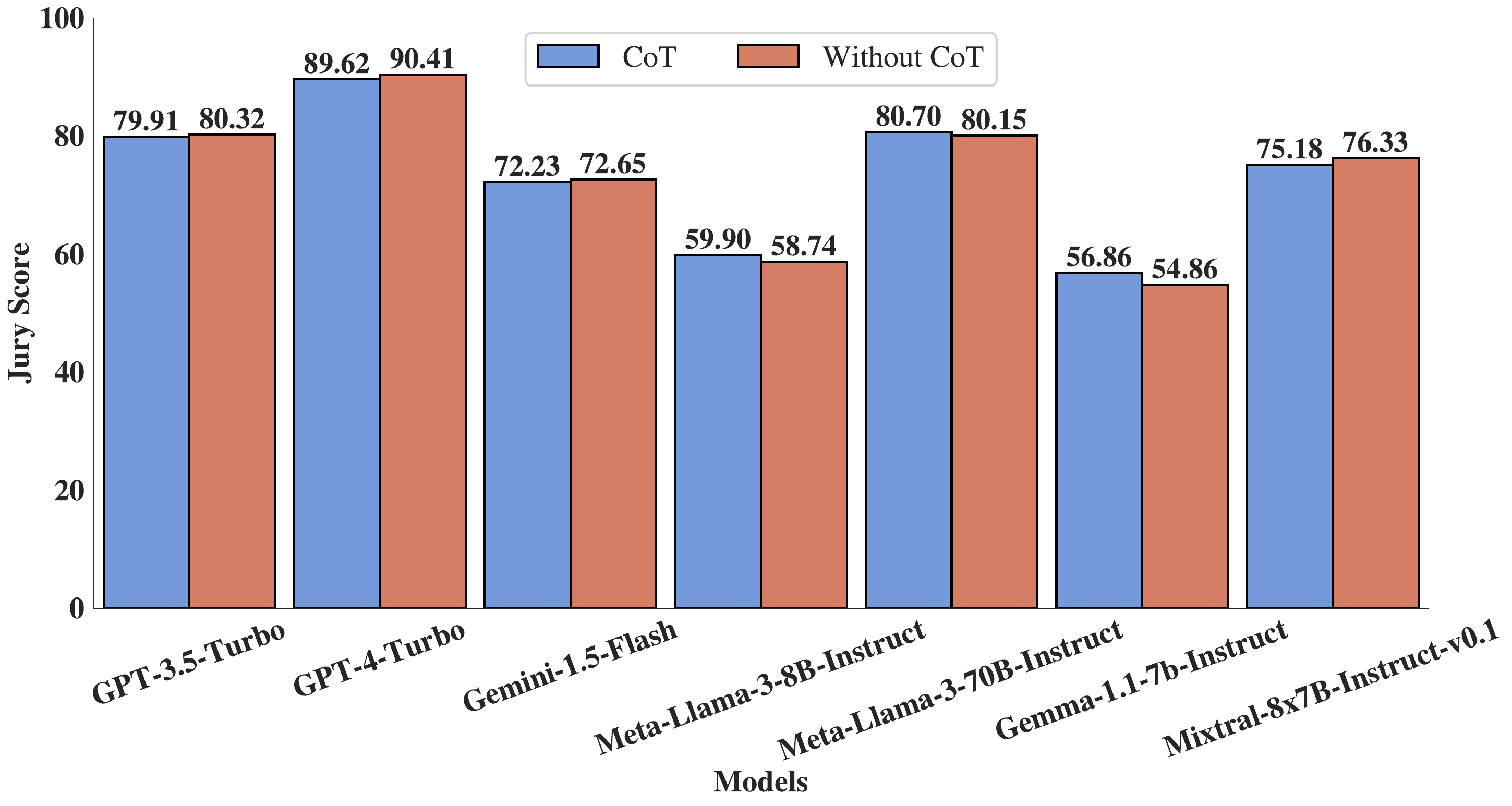}
        \label{fig:overall-pred}
        \caption{Rationale for the predicted answer.}
    \end{subfigure}
    \\
    \begin{subfigure}[c]{0.45\textwidth}
        \centering
        \includegraphics[width=\textwidth]{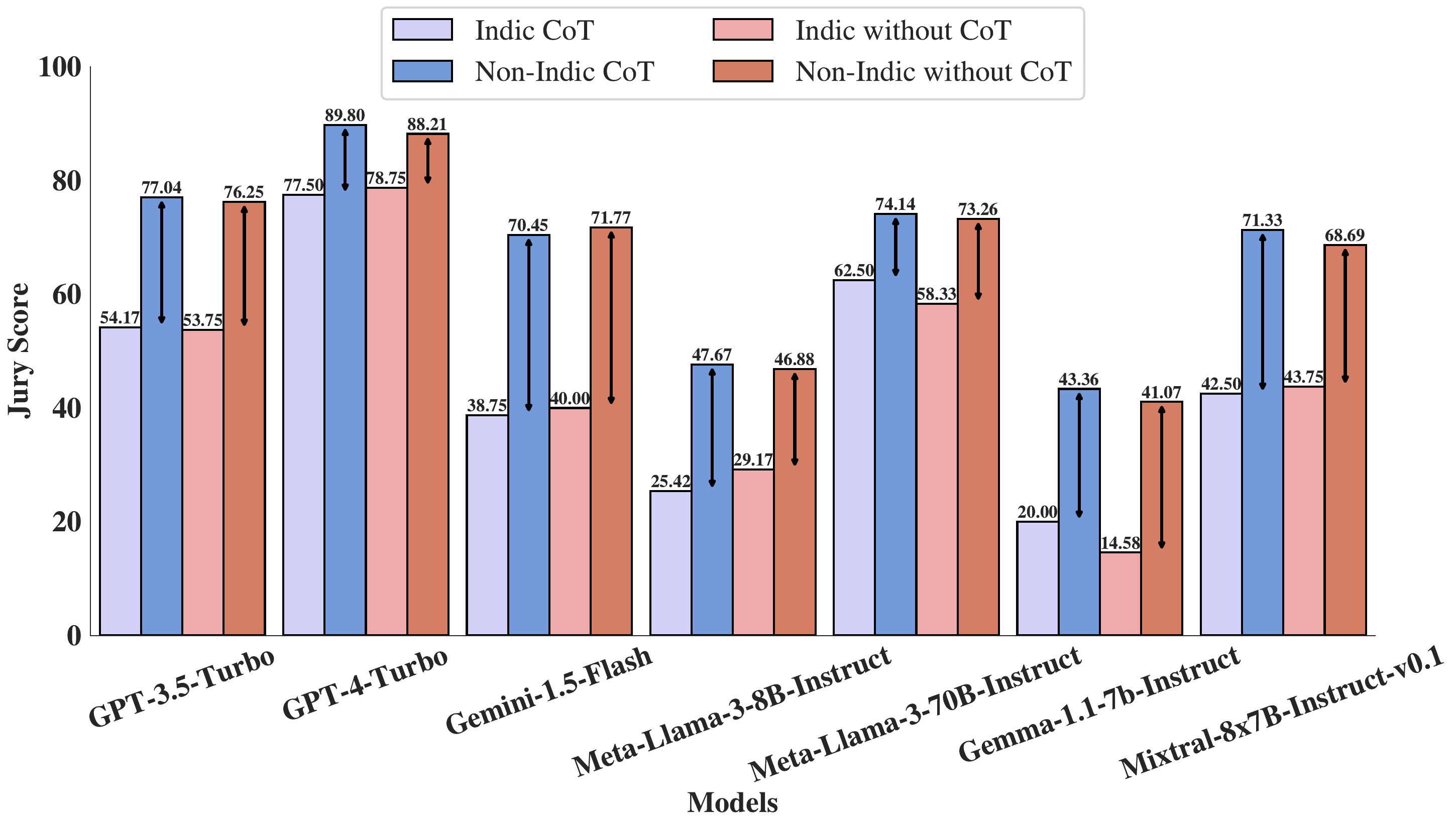}
        \caption{Predicted answer.}
        \label{fig:indic-correct}
    \end{subfigure}
    ~
    \begin{subfigure}[c]{0.45\textwidth}
        \centering
        \includegraphics[width=\textwidth]{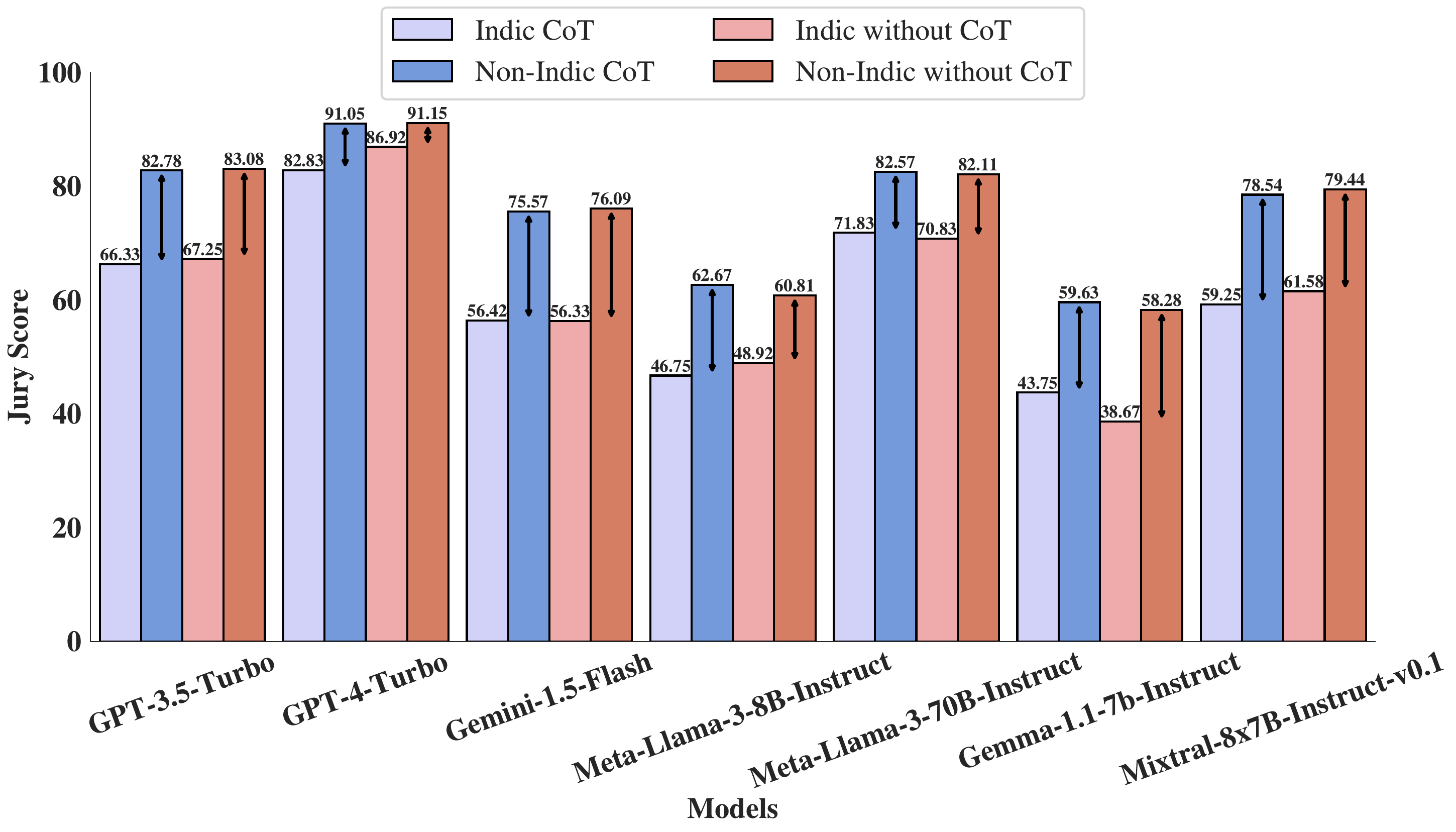}
        \caption{Rationale for the predicted answer.}
        \label{fig:indic-pred}
    \end{subfigure}
    \caption{Figures (a) and (b) display our aggregated results comparing scenarios with and without Chain-of-Thought (CoT) prompting. Figures (c) and (d) present a comparison between Indic and Non-Indic languages. The results are evaluated across three types of metrics: (i) the correctness of the predicted answer, and (ii) the correctness of the rationale for the prediction. All metrics are scaled from 0 to 100.}
    \label{fig:combined-plots}
    \vspace{-3mm}
\end{figure*}

%% file: result-benchmark.tex
\begin{table*}
\resizebox{\textwidth}{!}{
\begin{tabular}{l|l|rrrr|rrrr|rrrr}
\toprule
\multirow{2}{*}{\textbf{Model}} & \multirow{2}{*}{\textbf{Category}} & \multicolumn{4}{c|}{\textbf{Answer}} & \multicolumn{4}{c|}{\textbf{Rationales w/ Predicted Answer}} & \multicolumn{4}{c}{\textbf{Rationales w/ Gold Answer}} \\ \cline{3-14} 
 &  & \multicolumn{1}{c}{BLEU} & \multicolumn{1}{c}{Rouge} & \multicolumn{1}{c}{BS} & \multicolumn{1}{c|}{GEval} & \multicolumn{1}{c}{BLEU} & \multicolumn{1}{c}{Rouge} & \multicolumn{1}{c}{BS} & \multicolumn{1}{c|}{GEval} & \multicolumn{1}{c}{BLEU} & \multicolumn{1}{c}{Rouge} & \multicolumn{1}{c}{BS} & \multicolumn{1}{c}{GEval} \\ \midrule
\multirow{4}{*}{Gemini 1.5 Flash} & Indic & 31.3 & 72.8 & 91.7 & 40 & 6.6 & 29.2 & 86.8 & 56.2 & 19.7 & 30.6 & 87.6 & 87.6 \\
 & Non-Indic & 36.4 & 81.9 & 93.6 & 71 & 18.4 & 28 & 86.9 & 76 & 17.5 & 27.9 & 87.2 & 93.8 \\
 & All & 35.5 & 80.2 & 93.3 & 66 & 13.7 & 28.2 & 86.9 & 72.6 & 17.8 & 28.4 & 87.3 & 92.6 \\ \cdashline{2-14}
 & $\pm \mathit{\mathbf{\Delta}}(N-I)$ & $+5.1$ & $+9.1$ & $+1.9$ & $+31.0$ & $+11.8$ & $-1.2$ & $+0.1$ & $+19.8$ & $-2.2$ & $-2.7$ & $-0.4$ & $+6.2$ \\ \midrule
\multirow{4}{*}{Gemma-1.1-7B-it} & Indic & 13.2 & 45.7 & 89.2 & 14 & 19.4 & 28.5 & 86 & 38.6 & 22 & 30.3 & 87 & 85.2 \\
 & Non-Indic & 15.3 & 52.3 & 90.8 & 41 & 17.2 & 26 & 85.9 & 58.2 & 19.4 & 28.2 & 86.9 & 88.2 \\
 & All & 14.9 & 51.1 & 90.6 & 36 & 17.6 & 26.4 & 86 & 54.8 & 19.9 & 28.5 & 86.9 & 87.6 \\ \cdashline{2-14}
 & $\pm \mathit{\mathbf{\Delta}}(N-I)$ & $+2.1$ & $+6.6$ & $+1.6$ & $+27.0$ & $-2.2$ & $-2.5$ & $-0.1$ & $+19.6$ & $-2.6$ & $-2.1$ & $-0.1$ & $+3.0$ \\ \midrule
\multirow{4}{*}{GPT-3.5-Turbo} & Indic & 54.8 & 76.2 & 95 & 53 & 23.2 & 29.3 & 87.4 & 67.2 & 27.8 & 34.7 & 88.7 & 94 \\
 & Non-Indic & 52.8 & 81 & 95.9 & 76 & 22.3 & 29.1 & 87.6 & 83 & 23.9 & 31.4 & 88.2 & 94 \\
 & All & 53.1 & 80.1 & 95.8 & 72 & 22.5 & 29.1 & 87.5 & 80.2 & 24.6 & 32 & 88.3 & 94 \\ \cdashline{2-14}
 & $\pm \mathit{\mathbf{\Delta}}(N-I)$ & $-2.0$ & $+4.8$ & $+0.9$ & $+23.0$ & $-0.9$ & $-0.2$ & $+0.2$ & $+15.8$ & $-3.9$ & $-3.3$ & $-0.5$ & $0.0$ \\ \midrule
\multirow{4}{*}{GPT-4-Turbo} & Indic & 66.4 & 87.5 & 96.8 & 78 & 18.8 & 27.6 & 87.5 & 86.8 & 19 & 28.2 & 88.1 & 97 \\
 & Non-Indic & 65.5 & 90.4 & 97.4 & 88 & 17.1 & 25.6 & 87.3 & 91 & 16.8 & 27 & 87.6 & 97.6 \\
 & All & 65.7 & 89.8 & 97.3 & 86 & 17.4 & 26 & 87.3 & 90.4 & 17.1 & 27.2 & 87.7 & 97.6 \\ \cdashline{2-14}
 & $\pm \mathit{\mathbf{\Delta}}(N-I)$ & $-0.9$ & $+2.9$ & $+0.6$ & $+10.0$ & $-1.7$ & $-2.0$ & $-0.2$ & $+4.2$ & $-2.2$ & $-1.2$ & $-0.5$ & $+0.6$ \\ \midrule
\multirow{4}{*}{Meta-Llama-3-8B-Instruct} & Indic & 53.4 & 71.6 & 94 & 29 & 18.7 & 26.8 & 86.3 & 48.8 & 20.2 & 27.5 & 87.2 & 84.8 \\
 & Non-Indic & 54.2 & 75.1 & 95.3 & 46 & 18 & 25.2 & 86.4 & 60.8 & 19.4 & 27.4 & 87 & 84.6 \\
 & All & 54.1 & 74.4 & 95.1 & 43 & 18.1 & 25.5 & 86.3 & 58.6 & 19.5 & 27.4 & 87.1 & 84.6 \\ \cdashline{2-14}
 & $\pm \mathit{\mathbf{\Delta}}(N-I)$ & $+0.8$ & $+3.5$ & $+1.3$ & $+17.0$ & $-0.7$ & $-1.6$ & $+0.1$ & $+12.0$ & $-0.8$ & $-0.1$ & $-0.2$ & $-0.2$ \\ \midrule
\multirow{4}{*}{Meta-Llama-3-70B-Instruct} & Indic & 59.1 & 79.4 & 95.4 & 58 & 21.7 & 28.5 & 87 & 70.8 & 22.5 & 32.1 & 88.1 & 92.2 \\
 & Non-Indic & 61.7 & 84.2 & 96.7 & 73 & 21.3 & 28.7 & 87.3 & 82 & 21 & 29.2 & 87.6 & 94.2 \\
 & All & 61.2 & 83.3 & 96.5 & 70 & 21.4 & 28.7 & 87.3 & 80 & 21.2 & 29.7 & 87.7 & 93.8 \\ \cdashline{2-14}
 & $\pm \mathit{\mathbf{\Delta}}(N-I)$ & $+2.6$ & $+4.8$ & $+1.3$ & $+15.0$ & $-0.4$ & $+0.2$ & $+0.3$ & $+11.2$ & $-1.5$ & $-2.9$ & $-0.5$ & $+2.0$ \\ \midrule
\multirow{4}{*}{Mixtral-8x7B-Instruct-v0.1} & Indic & 4 & 21.5 & 86.1 & 43 & 16.8 & 26.3 & 87 & 61.4 & 16 & 25.6 & 87.2 & 93.4 \\
 & Non-Indic & 4.9 & 29.8 & 88 & 68 & 16.1 & 25.7 & 87 & 79.4 & 15.7 & 25.9 & 87.3 & 95.6 \\
 & All & 4.7 & 28.5 & 87.7 & 64 & 16.3 & 25.8 & 87 & 76.2 & 15.7 & 25.9 & 87.3 & 95.2 \\ \cdashline{2-14}
 & $\pm \mathit{\mathbf{\Delta}}(N-I)$ & $+0.9$ & $+8.3$ & $+1.9$ & $+25.0$ & $-0.7$ & $-0.6$ & $0.0$ & $+18.0$ & $-0.3$ & $+0.3$ & $+0.1$ & $+2.2$ \\
 \bottomrule

\end{tabular}
}
\caption{This table shows our complete evaluation results without using the Chain-of-Thought prompting technique. Here, $\mathit{\mathbf{\Delta}}(N-I)$ is the difference in performance between the Non-Indic and Indic subset. BS: BERTScore; GEval: Jury Evaluation.}
\label{tab:noncot-results-comb}
\end{table*}

\begin{table*}[]
\resizebox{\textwidth}{!}{
\begin{tabular}{l|l|rrrr|rrrr|rrrr}
\toprule
\multirow{2}{*}{\textbf{Model}} & \multirow{2}{*}{\textbf{Category}} & \multicolumn{4}{c|}{\textbf{Answer}} & \multicolumn{4}{c|}{\textbf{Rationales w/ Predicted Answer}} & \multicolumn{4}{c}{\textbf{Rationales w/ Gold Answer}} \\ \cline{3-14}
 &  & \multicolumn{1}{c}{BLEU} & \multicolumn{1}{c}{Rouge} & \multicolumn{1}{c}{BS} & \multicolumn{1}{c|}{GEval} & \multicolumn{1}{c}{BLEU} & \multicolumn{1}{c}{Rouge} & \multicolumn{1}{c}{BS} & \multicolumn{1}{c|}{GEval} & \multicolumn{1}{c}{BLEU} & \multicolumn{1}{c}{Rouge} & \multicolumn{1}{c}{BS} & \multicolumn{1}{c}{GEval} \\ \midrule
\multirow{4}{*}{Gemini 1.5 Flash} & Indic & 1.3 & 72.6 & 91.3 & 38 & 4.7 & 28.2 & 86.7 & 56.4 & 18.9 & 29.2 & 87.4 & 92.4 \\
 & Non-Indic & 34.8 & 81.7 & 93.7 & 70 & 18.5 & 28.3 & 87.1 & 75.4 & 17.7 & 27.9 & 87.2 & 93.6 \\
 & All & 6.9 & 80.1 & 93.3 & 64 & 12.1 & 28.3 & 87.1 & 72.2 & 17.9 & 28.1 & 87.2 & 93.4 \\ \cdashline{2-14}
 & $\pm \mathit{\mathbf{\Delta}}(N-I)$ & $+33.5$ & $+9.1$ & $+2.4$ & $+32.0$ & $+13.8$ & $+0.1$ & $+0.4$ & $+19.0$ & $-1.2$ & $-1.3$ & $-0.2$ & $+1.2$ \\ \midrule
\multirow{4}{*}{Gemma-1.1-7B-it} & Indic & 8.4 & 31.9 & 87.5 & 20 & 20.2 & 30.5 & 86.2 & 43.6 & 19.8 & 29.3 & 86.7 & 87 \\
 & Non-Indic & 10.1 & 39.9 & 89.3 & 43 & 17.2 & 26.5 & 86.1 & 59.6 & 18 & 27.4 & 86.6 & 88.4 \\
 & All & 9.8 & 38.6 & 89 & 39 & 17.8 & 27.2 & 86.1 & 56.8 & 18.3 & 27.7 & 86.6 & 88.2 \\ \cdashline{2-14}
 & $\pm \mathit{\mathbf{\Delta}}(N-I)$ & $+1.7$ & $+8.0$ & $+1.8$ & $+23.0$ & $-3.0$ & $-4.0$ & $-0.1$ & $+16.0$ & $-1.8$ & $-1.9$ & $-0.1$ & $+1.4$ \\ \midrule
\multirow{4}{*}{GPT-3.5-Turbo} & Indic & 44.8 & 74 & 94.6 & 54 & 24.7 & 30.2 & 87.6 & 66.2 & 28 & 34.7 & 88.7 & 94.8 \\
 & Non-Indic & 53.7 & 82.2 & 96 & 77 & 22.3 & 29.6 & 87.6 & 82.6 & 23.1 & 31.4 & 88 & 95.6 \\
 & All & 52 & 80.7 & 95.8 & 73 & 22.7 & 29.7 & 87.6 & 79.8 & 23.9 & 32 & 88.2 & 95.4 \\ \cdashline{2-14}
 & $\pm \mathit{\mathbf{\Delta}}(N-I)$ & $+8.9$ & $+8.2$ & $+1.4$ & $+23.0$ & $-2.4$ & $-0.6$ & $0.0$ & $+16.4$ & $-4.9$ & $-3.3$ & $-0.7$ & $+0.8$ \\ \midrule
\multirow{4}{*}{GPT-4-Turbo} & Indic & 67.1 & 87.3 & 97 & 77 & 18.5 & 27.2 & 87.5 & 82.8 & 18.9 & 29.6 & 88.2 & 97.6 \\
 & Non-Indic & 66.6 & 90.4 & 97.5 & 89 & 16.6 & 26 & 87.3 & 91 & 16.6 & 26.5 & 87.5 & 97.6 \\
 & All & 66.6 & 89.9 & 97.4 & 87 & 16.9 & 26.2 & 87.3 & 89.6 & 17 & 27 & 87.6 & 97.6 \\ \cdashline{2-14}
 & $\pm \mathit{\mathbf{\Delta}}(N-I)$ & $-0.5$ & $+3.1$ & $+0.5$ & $+12.0$ & $-1.9$ & $-1.2$ & $-0.2$ & $+8.2$ & $-2.3$ & $-3.1$ & $-0.7$ & $0.0$ \\ \midrule
\multirow{4}{*}{Meta-Llama-3-8B-Instruct} & Indic & 48.6 & 67.4 & 93.3 & 25 & 19.7 & 27.7 & 86.5 & 46.6 & 20.7 & 29 & 87.2 & 77 \\
 & Non-Indic & 55.3 & 75.7 & 95.4 & 47 & 17.6 & 25.1 & 86.4 & 62.6 & 19.8 & 27.9 & 87.1 & 82.8 \\
 & All & 54.1 & 74.2 & 95 & 43 & 17.9 & 25.6 & 86.4 & 59.8 & 20 & 28.1 & 87.1 & 81.8 \\ \cdashline{2-14}
 & $\pm \mathit{\mathbf{\Delta}}(N-I)$ & $+6.7$ & $+8.3$ & $+2.1$ & $+22.0$ & $-2.1$ & $-2.6$ & $-0.1$ & $+16.0$ & $-0.9$ & $-1.1$ & $-0.1$ & $+5.8$ \\ \midrule
\multirow{4}{*}{Meta-Llama-3-70B-Instruct} & Indic & 59.5 & 81 & 95.7 & 62 & 21.6 & 28.3 & 87 & 71.8 & 24.9 & 32.2 & 88.2 & 95.6 \\
 & Non-Indic & 62 & 84.1 & 96.7 & 74 & 21.2 & 28.8 & 87.3 & 82.4 & 21.5 & 30 & 87.8 & 94.2 \\
 & All & 61.5 & 83.6 & 96.5 & 72 & 21.3 & 28.7 & 87.2 & 80.6 & 22.1 & 30.4 & 87.9 & 94.4 \\ \cdashline{2-14}
 & $\pm \mathit{\mathbf{\Delta}}(N-I)$ & $+2.5$ & $+3.1$ & $+1.0$ & $+12.0$ & $-0.4$ & $+0.5$ & $+0.3$ & $+10.6$ & $-3.4$ & $-2.2$ & $-0.4$ & $-1.4$ \\ \midrule
\multirow{4}{*}{Mixtral-8x7B-Instruct-v0.1} & Indic & 3.8 & 21.4 & 86 & 42 & 16.8 & 27.1 & 86.9 & 59.2 & 16.8 & 26.2 & 87.6 & 94 \\
 & Non-Indic & 4.9 & 32.4 & 88.4 & 71 & 16.1 & 25.8 & 87.1 & 78.4 & 15.7 & 26.1 & 87.4 & 95 \\
 & All & 4.7 & 30.5 & 88 & 66 & 16 & 26.1 & 87 & 75 & 15.7 & 26.2 & 87.4 & 94.8 \\ \cdashline{2-14}
 & $\pm \mathit{\mathbf{\Delta}}(N-I)$ & $+1.1$ & $+11.0$ & $+2.4$ & $+29.0$ & $-0.7$ & $-1.3$ & $+0.2$ & $+19.2$ & $-1.1$ & $-0.1$ & $-0.2$ & $+1.0$ \\ \bottomrule
\end{tabular}
}
\caption{This table shows our complete evaluation results using the Chain-of-Thought prompting technique. Here, $\mathit{\mathbf{\Delta}}(N-I)$ is the difference in performance between the Non-Indic and Indic subset. BS: BERTScore; GEval: Jury Evaluation.}
\label{tab:cot-results-comb}
\end{table*}